\documentclass[twoside,11pt,letterpaper]{article}
\input epsf %% at top of file
\usepackage[active]{srcltx}
\usepackage{jair}
\usepackage{theapa}

\usepackage[T1]{fontenc}
\usepackage[latin1]{inputenc}
\usepackage{subfigure}
\usepackage{amsmath}
\usepackage{graphicx}
\usepackage{amssymb}
\usepackage{algorithm}
\usepackage[noend]{algorithmic}
\usepackage{color}

\newtheorem{proposition}{Proposition}

\jairheading{25}{2006}{349-387}{06/05}{03/06} \ShortHeadings{Representing Conversations for Scalable
Overhearing} {Gutnik \& Kaminka} \firstpageno{349}

\input epsf %% at top of file

\begin{document}

%\title{A Petri Net Representation of Multi-Agent Conversations
%for Scalable Overhearing}

\title{Representing Conversations for Scalable Overhearing}

\author{\name Gery Gutnik \email gutnikg@cs.biu.ac.il \\
       \name Gal A. Kaminka \email galk@cs.biu.ac.il \\
       \addr Computer Science Department \\
       Bar Ilan University \\ Ramat Gan 52900, Israel
       }

% For research notes, remove the comment character in the line below.
% \researchnote

\maketitle

\begin{abstract}
Open distributed multi-agent systems are gaining interest in the academic community and in industry. In such
open settings, agents are often coordinated using standardized agent conversation protocols. The representation
of such protocols (for analysis, validation, monitoring, etc) is an important aspect of multi-agent
applications. Recently, Petri nets have been shown to be an interesting approach to such representation, and
radically different approaches using Petri nets have been proposed. However, their relative strengths and
weaknesses have not been examined. Moreover, their scalability and suitability for different tasks have not been
addressed.  This paper addresses both these challenges. First, we analyze existing Petri net representations in
terms of their scalability and appropriateness for overhearing, an important task in monitoring open multi-agent
systems. Then, building on the insights gained, we introduce a novel representation using Colored Petri nets
that explicitly represent legal joint conversation states and messages. This representation approach offers
significant improvements in scalability and is particularly suitable for overhearing. Furthermore, we show that
this new representation offers a comprehensive coverage of all conversation features of FIPA conversation
standards. We also present a procedure for transforming AUML conversation protocol diagrams (a standard
human-readable representation), to our Colored Petri net representation.
\end{abstract}

%-------------------------------------------------------------------------
\section{Introduction}
\label{Introduction}

Open distributed multi-agent systems (MAS) are composed of multiple, independently-built agents that carry out
mutually-dependent tasks. In order to allow inter-operability of agents of different designs and implementation,
the agents often coordinate using standardized interaction protocols, or conversations. Indeed, the multi-agent
community has been investing a significant effort in developing standardized \emph{Agent Communication
Languages} (ACL) to facilitate sophisticated multi-agent systems \cite{finin97,kone00,chaib02,fipa}. Such
standards define communicative acts, and on top of them, \emph{interaction protocols}, ranging from simple
queries as to the state of another agent, to complex negotiations by auctions or bidding on contracts. For
instance, the \emph{FIPA Contract Net Interaction Protocol} \cite{fipac} defines a concrete set of message
sequences that allows the interacting agents to use the contract net protocol for negotiations.

Various formalisms have been proposed to describe such standards \cite<e.g.,
>{smith96,parunak96,odell00,odell01a,auml}. In particular,
AUML--\emph{Agent Unified Modelling Language}--is currently used in the FIPA-ACL standards
\cite{fipad,fipac,fipaa,fipab,odell01b} \footnote{\cite{fipaa} is currently deprecated. However, we use this
specification since it describes many important features needed in modelling multi-agent interactions.}. UML 2.0
\cite{auml}, a new emerging standard influenced by AUML, has the potential to become the FIPA-ACL standard (and
a forthcoming IEEE standard) in the future. However, for the moment, a large set of FIPA specifications remains
formalized using AUML. While AUML is intended for human readability and visualization, interaction protocols
should ideally be represented in a way that is amenable to automated analysis, validation and verification,
online monitoring, etc.

Lately, there is increasing interest in using Petri nets \cite{petriNets} in modelling multi-agent interaction
protocols
\cite{cost99a,cost99b,cost00,lin00,nowostawski01,purvis02,cranefield02,ramos02,mazouzi02,poutakidis02}. There is
broad literature on using Petri nets to analyze the various aspects of distributed systems \cite<e.g. in
deadlock detection as shown by >{khomenko00}, and there has been recent work on specific uses of Petri nets in
multi-agent systems, e.g., in validation and testing \cite{desel97}, in automated debugging and monitoring
\cite{poutakidis02}, in dynamic interpretation of interaction protocols \cite{cranefield02,desilva03}, in
modelling agents behavior induced by their participation in a conversation \cite{ling03} and in interaction
protocols refinement allowing modular construction of complex conversations \cite{hameurlain03}.

However, key questions remain open on the use of Petri nets for conversation representation. First, while
radically different approaches to representation using Petri nets have been proposed, their relative strengths
and weaknesses have not been investigated. Second, many investigations have only addressed restricted subsets of
the features needed in representing complex conversations such as those standardized by FIPA (see detailed
discussion of previous work in Section~\ref{reprentation}). Finally, no procedures have been proposed for
translating human-readable AUML protocol descriptions into the corresponding machine-readable Petri nets.

This paper addresses these open challenges in the context of scalable \emph{overhearing}. Here, an overhearing
agent passively tracks many concurrent conversations involving multiple participants, based solely on their
exchanged messages, while not being a participant in any of the overheard conversations itself
\cite{novick93,busetta01,kaminka02,poutakidis02,busetta02,legras02,gutnik04,rossi04}. Overhearing is useful in
visualization and progress monitoring \cite{kaminka02}, in detecting failures in interactions
\cite{poutakidis02}, in maintaining organizational and situational awareness \cite{novick93,legras02,rossi04}
and in non-obtrusively identifying opportunities for offering assistance \cite{busetta01,busetta02}. For
instance, an overhearing agent may monitor the conversation of a contractor agent engaged in multiple
contract-net protocols with different bidders and bid callers, in order to detect failures.

We begin with an analysis of Petri net representations, with respect to scalability and overhearing. We classify
representation choices along two dimensions affecting scalability: (i) the technique used to represent multiple
concurrent conversations; and (ii) the choice of representing either individual or joint interaction states. We
show that while the run-time complexity of monitoring conversations using different approaches is the same,
choices along these two dimensions have significantly different space requirements, and thus some choices are
more scalable (in the number of conversations) than others. We also argue that representations suitable for
overhearing require the use of explicit message places, though only a subset of previously-explored techniques
utilized those.

Building on the insights gained, the paper presents a novel representation that uses Colored Petri nets
(CP-nets) in which places explicitly denote messages, and valid joint conversation states. This representation
is particularly suited for overhearing as the number of conversations is scaled-up. We show how this
representation can be used to represent essentially all features of FIPA AUML conversation standards, including
simple and complex interaction building blocks, communicative act attributes such as message guards and
cardinalities, nesting, and temporal aspects such as deadlines and duration.

To realize the advantages of machine-readable representations, such as for debugging \cite{poutakidis02},
existing human-readable protocol descriptions must be converted to their corresponding Petri net
representations. As a final contribution in this paper, we provide a skeleton semi-automated procedure for
converting FIPA conversation protocols in AUML to Petri nets, and demonstrate its use on a complex FIPA
protocol. While this procedure is not fully automated, it takes a first step towards addressing this open
challenge.

This paper is organized as follows. Section~\ref{reprentation} presents the motivation for our work. Sections
\ref{simple-and-complex-building-blocks} through \ref{temporal-aspects} then present the proposed representation
addressing all FIPA conversation features including basic interaction building blocks
(Section~\ref{simple-and-complex-building-blocks}), message attributes (Section~\ref{interaction-attributes}),
nested \& interleaved interactions (Section~\ref{nested-and-interleaved-interactions}), and temporal aspects
(Section~\ref{temporal-aspects}). Section~\ref{algorithm} ties these features together: It presents a skeleton
algorithm for transforming an AUML protocol diagram to its Petri net representation, and demonstrates its use on
a challenging FIPA conversation protocol. Section~\ref{summary} concludes. The paper rounds up with three
appendixes. The first provides a quick review of Petri nets. Then, to complete coverage of FIPA interactions,
Appendix~\ref{additional-examples} provides additional interaction building blocks.
Appendix~\ref{complex-example} presents a Petri net of a complex conversation protocol, which integrates many of
the features of the developed representation technique.

%-------------------------------------------------------------------------
\section{Representations for Scalable Overhearing}
\label{reprentation}

Overhearing involves monitoring conversations as they progress, by tracking messages that are exchanged between
participants \cite{gutnik04}. We are interested in representations that can facilitate \emph{scalable}
overhearing, tracking many concurrent conversations, between many agents. We focus on open settings, where the
complex internal state and control logic of agents is not known in advance, and therefore exclude discussions of
Petri net representations which explicitly model agent internals \cite<e.g.,>{moldt97,xu01}.  Instead, we treat
agents as black boxes, and consider representations that commit only to the agent's conversation state (i.e.,
its role and progress in the conversation).

The suitability of a representation for scalable overhearing is affected by several facets. First, since
overhearing is based on tracking messages, the representation must be able to explicitly represent the passing
of a message (communicative act) from one agent to another (Section~\ref{rep-message-monitoring}). Second, the
representation must facilitate tracking of multiple concurrent conversations. While the tracking runtime is
bounded from below by the number of messages (since in any case, all messages are overheard and processed),
space requirements may differ significantly (see
Sections~\ref{rep-single-conversation}--\ref{rep-multiple-conversations}).

%-------------------------------------------------------------------------
\subsection{Message-monitoring versus state-monitoring}
\label{rep-message-monitoring}

We distinguish two settings for tracking the progress of conversations, depending on the information available
to the tracking agent. In the first type of setting, which we refer to as \emph{state monitoring}, the tracking
agent has access to the \emph{internal} state of the conversation in one or more of the participants, but not
necessarily to the messages being exchanged. The other settings involves \emph{message monitoring}, where the
tracking agent has access only to the messages being exchanged (which are \emph{externally observable}), but
cannot directly observe the internal state of the conversation in each participant. Overhearing is a form of
message monitoring.

Representations that support state monitoring use places to denote the conversation states of the participants.
Tokens placed in these places (the net marking) denote the current state. The sending or receiving of a message
by a participant is not explicitly represented, and is instead implied by moving tokens (through transition
firings) to the new state places. Thus, such a representation essentially assumes that the internal conversation
state of participants is directly observable by the monitoring agent. Previous work utilizing state monitoring
includes work by \citeA{cost99a,cost99b,cost00,lin00,mazouzi02,ramos02}.

The representation we present in this paper is intended for overhearing tasks, and cannot assume that the
conversation states of overheard agents are observable. Instead, it must support message monitoring, where in
addition to using tokens in state places (to denote current conversation state), the representation uses
\emph{message places}, where tokens are placed when a corresponding message is overheard. A conversation-state
place and a message place are connected via a transition to a state place denoting the new conversation state.
Tokens placed in these originating places--indicating a message was received at an appropriate conversation
state--will cause the transition to fire, and for the tokens to be placed in the new conversation state place.
Thus the new conversation state is inferred from "observing" a message. Previous investigations, that have used
explicit message places, include work by
\citeA{cost99a,cost99b,cost00,nowostawski01,purvis02,cranefield02,poutakidis02}\footnote{\citeA{cost99a,cost99b,cost00}
present examples of both state- and message- monitoring representations.}. These are discussed in depth below.

%-------------------------------------------------------------------------
\subsection{Representing a Single Conversation}
\label{rep-single-conversation}

Two representation variants are popular within those that utilize conversation places (in addition to message
places): \emph{Individual state representations} use separate places and tokens for the state of each
participant (each role). Thus, the overall state of the conversation is represented by different tokens marking
multiple places. \emph{Joint state representations} use a single place for each joint conversation state of all
participants. The placement of a token within such a place represents the overhearing agent's belief that the
participants are in the appropriate joint state.

Most previous representations use individual states. In these, different markings distinguish a conversation
state where one agent has sent a message, from a state where the other agent received it. The net for each
conversation role is essentially built separately, and is merged with the other nets, or connected to them via
fusion places or similar means.

\citeA{cost99a,cost99b,cost00} have used CP-nets with individual state places for representing KQML and FIPA
interaction protocols. Transitions represent message events, and CP-net features, such as token colors and arc
expressions, are used to represent AUML message attributes and sequence expressions. The authors also point out
that deadlines (a temporal aspect of interaction) can be modelled, but no implementation details are provided.
\citeA{cost99a} also proposed using hierarchical CP-nets to represent hierarchical multi-agent conversations.

\citeA{purvis02,cranefield02} represented conversation roles as separate CP-nets, where places denote both
interaction messages and states, while transitions represent operations performed on the corresponding
communicative acts such as send, receive, and process. Special in/out places are used to pass net tokens between
the different CP-nets, through special get/put transitions, simulating the actual transmission of the
corresponding communicative acts.

In principle, individual-state representations require two places in each role, for every message. For a given
message, there would be two individual places for the sender (before sending and after sending), and similarly
two more \emph{for each receiver} (before receiving and after receiving). All possible conversation
states--valid or not--can be represented. %by different markings.
For a single message and two roles, there are two places for each role (four places total), and four possible
conversation states: message sent and received, sent and not received, not sent but incorrectly believed to have
been received, not sent and not received. These states can be represented by different markings. For instance, a
conversation state where the message has been sent but not received is denoted by a token in the
\emph{'after-sending'} place of the \emph{sender} and another token in the \emph{'before-receiving'} place of
the \emph{receiver}. This is summarized in the following proposition:

\begin{proposition}\label{prop-indv}Given a conversation with $R$
roles and a total of $M$ possible messages, an individual state representation has space complexity of $O(MR)$.
\end{proposition}

While the representations above all represent each role's conversation state separately, many applications of
overhearing only require representation of valid conversation states (message not sent and not received, or sent
and received). Indeed, specifications for interaction protocols often assume the use of underlying
synchronization protocols to guarantee delivery of messages \cite{paurobally03a,paurobally03b}. Under such an
assumption, for every message, there are only two joint states regardless of the number of roles. For example,
for a single message and three roles--a sender and two receivers, there are two places and two possible
markings: A token in a \emph{before sending/receiving} place represents a conversation state where the message
has not yet been sent by the \emph{sender} (and the two \emph{receivers} are waiting for it), while a token in a
\emph{after sending/receiving} place denotes that the message has been sent and received by both
\emph{receivers}.

\citeA{nowostawski01} utilize CP-nets where places denote joint conversation states. They also utilize places
representing communicative acts. \citeA{poutakidis02} proposed a representation based on Place-Transition nets
(PT-nets)--a more restricted representation of Petri nets that has no color. They presented several interaction
building blocks, which could then fit together to model additional conversation protocols.  In general, the
following proposition holds with respect to such representations:

\begin{proposition}\label{prop-joint}{Given a conversation with $R$
roles and a total of $M$ possible messages, a joint state representation that represents only legal states has
space complexity of $O(M)$.}\end{proposition}

The condition of representing only \emph{valid} states is critical to the complexity analysis.  If all joint
conversation states--valid and invalid--are to be represented, the space complexity would be $O(M^{R})$.  In
such a case, an individual-state representation would have an advantage. This would be the case, for instance,
if we do not assume the use of synchronization protocols, e.g., where the overhearing agent may wish to track
the exact system state even while a message is underway (i.e., sent and not yet received).

%-------------------------------------------------------------------------
\subsection{Representing Multiple Concurrent Conversations}
\label{rep-multiple-conversations}

Propositions~\ref{prop-indv} and \ref{prop-joint} above address the space complexity of representing a single
conversation. However, in large scale systems an overhearing agent may be required to monitor multiple
conversations in parallel. For instance, an overhearing agent may be monitoring a middle agent that is carrying
multiple parallel instances of a single interaction protocol with multiple partners, e.g., brokering
\cite{fipad}.

Some previous investigations propose to duplicate the appropriate Petri net representation for each monitored
conversation \cite{nowostawski01,poutakidis02}. In this approach, every conversation is tracked by a separate
Petri-net, and thus the number of Petri nets (and their associated tokens) grows with the number of
conversations (Proposition~\ref{prop-pt-nets}).  For instance, \citeA{nowostawski01} shows an example where a
contract-net protocol is carried out with three different contractors, using three duplicate CP-nets.  This is
captured in the following proposition:

\begin{proposition}\label{prop-pt-nets} A representation that creates
multiple instances of a conversation Petri net to represent $C$ conversations, requires $O(C)$ net structures,
and $O(C)$ bits for all tokens. \end{proposition}

Other investigations take a different approach, in which a single CP-net structure is used to monitor all
conversations of the same protocol. The tokens associated with conversations are differentiated by their token
color \cite{cost99a,cost99b,cost00,lin00,mazouzi02,cranefield02,purvis02,ramos02}. For example, by assigning
each token a color of the tuple type $\langle sender, receiver\rangle$, an agent can differentiate multiple
tokens in the same place and thus track conversations of different pairs of agents\footnote{See
Section~\ref{interaction-attributes} to distinguish between different conversations by the same agents.}. Color
tokens use multiple bits per token; up to $\log{C}$ bits are required to differentiate $C$ conversations.
Therefore, the number of bits required to track $C$ conversations using $C$ tokens is $C\log{C}$. This leads to
the following proposition.

\begin{proposition}\label{prop-cp-nets}A representation that uses
color tokens to represent $C$ multiple instances of a conversation, requires $O(1)$ net structures, and
$O(C\log{C})$ bits for all tokens. \end{proposition}

Due to the constants involved, the space requirements of Proposition~\ref{prop-pt-nets} are in practice much
more expensive than those of Proposition~\ref{prop-cp-nets}. Proposition~\ref{prop-pt-nets} refers to the
creation of $O(C)$ Petri networks, each with duplicated place and transition data structures. In contrast,
Proposition~\ref{prop-cp-nets} refers to bits required for representing $C$ color tokens on a single CP net.
Moreover, in most practical settings, a sufficiently large constant bound on the number of conversations may be
found, which will essentially reduce the $O(\log C)$ factor to $O(1)$.

Based on Propositions~\ref{prop-indv}--\ref{prop-cp-nets}, it is possible to make concrete predictions as to the
scalability of different approaches with respect to the number of conversations, roles.
Table~\ref{scalability-comparison} shows the space complexity of different approaches when modelling $C$
conversations of the same protocol, each with a maximum of $R$ roles, and $M$ messages, under the assumption of
underlying synchronization protocols. The table also cites relevant previous work.

\begin{table} [ht]
\begin{center}
\begin{tabular}{|c|c|c|} \hline\hline
  & \multicolumn{2}{c|}{\textbf{Representing Multiple Conversations (of
Same  Protocol)}}\\ \hline
  & \textbf{Multiple CP- or PT-nets} & \textbf{Using color tokens,
single CP-net} \\
  & (Proposition~\ref{prop-pt-nets}) & (Proposition~\ref{prop-cp-nets})
\\  \hline

  &  & \textbf{Space: $O(MR+C\log{C})$} \\
  \textbf{Individual} & & \citeA{cost99a,cost99b,cost00},\\
  \textbf{States} & \textbf{Space: $O(MRC)$} &
\citeA{lin00,cranefield02},\\
  (Proposition~\ref{prop-indv}) & & \citeA{purvis02,ramos02},\\
  & & \citeA{mazouzi02} \\ \hline
 % & &  \\ \hline
 % & & \\
  \textbf{Joint} & \textbf{Space: $O(MC)$} & \textbf{Space:
$O(M+C\log{C})$} \\
  \textbf{States} & \citeA{nowostawski01}, & This paper\\
  (Proposition~\ref{prop-joint}) & \citeA{poutakidis02}& \\
%  & & \\
  \hline\hline
\end{tabular}
\vspace{-5pt}
\end{center}
\caption{Scalability of different representations} \label{scalability-comparison}
\end{table}

Building on the insights gained from Table~\ref{scalability-comparison}, we propose a representation using
CP-nets where places explicitly represent joint conversation states (corresponding to the lower-right cell in
Table~\ref{scalability-comparison}), and tokens color is used to distinguish concurrent conversations (as in the
upper-right cell in Table \ref{scalability-comparison}). As such, it is related to the works that have these
features, but as the table demonstrates, is a novel synthesis.

Our representation uses similar structures to those found in the works of \citeA{nowostawski01} and
\citeA{poutakidis02}. However, in contrast to these previous investigations, we rely on token color in CP-nets
to model concurrent conversations, with space complexity $O(M+C\log{C})$. We also show
(Sections~\ref{simple-and-complex-building-blocks}--\ref{temporal-aspects}) how it can be used to cover a
variety of conversation features not covered by these investigations. These features include representation of a
full set of FIPA interaction building blocks, communicative act attributes (such as message guards, sequence
expressions, etc.), compact modelling of concurrent conversations, nested and interleaved interactions, and
temporal aspects.

%-------------------------------------------------------------------------
\section{Representing Simple \& Complex Interaction Building Blocks}
\label{simple-and-complex-building-blocks}

\begin{figure}[ht]
\begin{center}
\subfigure[AUML representation]{
\includegraphics[width=0.35\columnwidth,keepaspectratio]{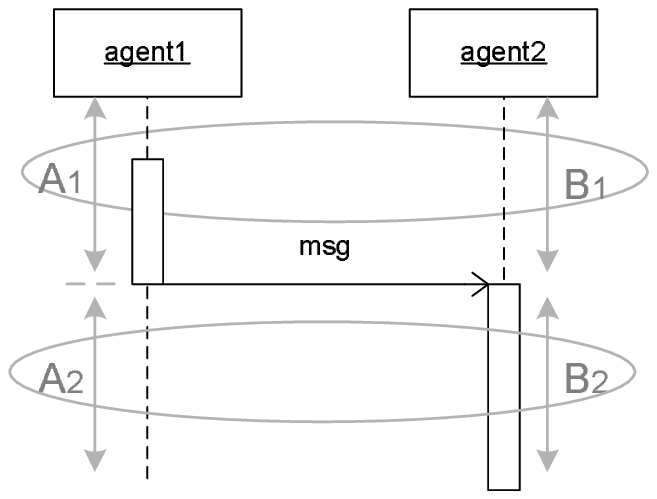}}
\subfigure[CP-net representation]{
\includegraphics[width=0.61\columnwidth,keepaspectratio]{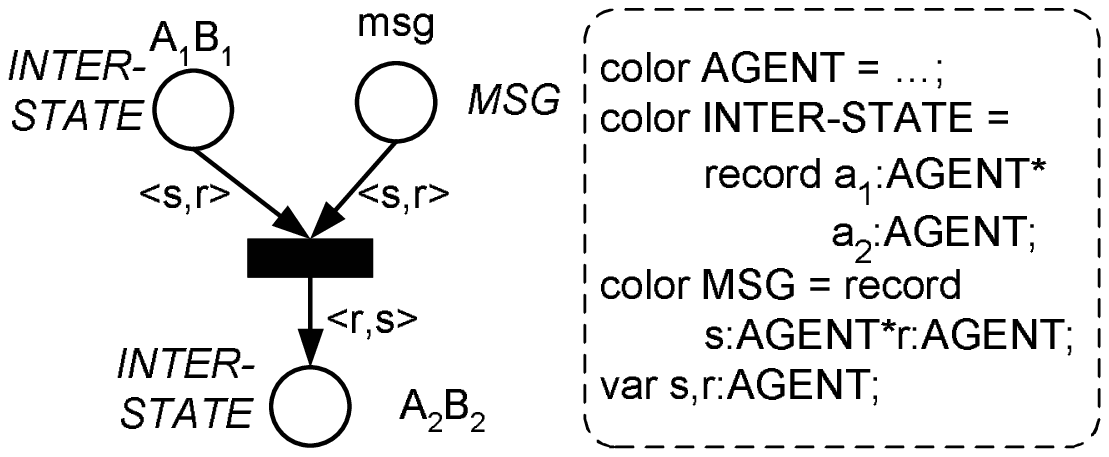}}
\caption {Asynchronous message interaction.}
  \label{fig-asynch-msg}
\vspace{-11pt}
  \end{center}
\end{figure}

This section introduces the fundamentals of our representation, and demonstrates how various simple and complex
AUML interaction messages, used in FIPA conversation standards \cite{fipaa}, can be implemented using the
proposed CP-net representation. We begin with a simple conversation, shown in Figure~\ref{fig-asynch-msg}-a
using an AUML protocol diagram. Here, $agent_{1}$ sends an asynchronous message $msg$ to $agent_{2}$.

To represent agent conversation protocols, we define two types of places, corresponding to messages and
conversation states. The first type of net places, called \emph{message places}, is used to describe
conversation communicative acts. Tokens placed in message places indicate that the associated communicative act
has been overheard.  The second type of net places, \emph{agent places}, is associated with the valid
\emph{joint} conversation states of the interacting agents. Tokens placed in agent places indicate the current
joint state of the conversation within the interaction protocol.

Transitions represent the transmission and receipt of communicative acts between agents.  Assuming underlying
synchronization protocols, a transition always originates within a joint-state place and a message place, and
targets a joint conversation state (more than one is possible--see below). Normally, the current conversation
state is known (marked with a token), and must wait the overhearing of the matching message (denoted with a
token at the connected message place). When this token is marked, the transition fires, automatically marking
the new conversation state.

Figure~\ref{fig-asynch-msg}-b presents CP net representation of the earlier example of
Figure~\ref{fig-asynch-msg}-a. The CP-net in Figure~\ref{fig-asynch-msg}-b has three places and one transition
connecting them. The $A_{1}B_{1}$ and the $A_{2}B_{2}$ places are agent places, while the $msg$ place is a
message place. The $A$ and $B$ capital letters are used to denote the $agent_{1}$ and the $agent_{2}$ individual
    interaction states respectively (we have indicated the individual and
the joint interaction states over the AUML diagram in Figure~\ref{fig-asynch-msg}-a, but omit these annotations
in later figures). Thus, the $A_{1}B_{1}$ place indicates a joint interaction state where $agent_{1}$ is ready
to send the $msg$ communicative act to $agent_{2}$ ($A_{1}$) and $agent_{2}$ is waiting to receive the
corresponding message ($B_{1}$). The $msg$ message place corresponds to the $msg$ communicative act sent between
the two agents. Thus, the transmission of the $msg$ communicative act causes the agents to transition to the
$A_{2}B_{2}$ place. This place corresponds to the joint interaction state in which $agent_{1}$ has already sent
the $msg$ communicative act to $agent_{2}$ ($A_{2}$) and $agent_{2}$ has received it ($B_{2}$).

The CP-net implementation in Figure~\ref{fig-asynch-msg}-b also introduces the use of token colors to represent
additional information about interaction states and communicative acts. The token color sets are defined in the
net declaration, i.e. the dashed box in Figure~\ref{fig-asynch-msg}-b. The syntax follows the standard CPN ML
notation \cite{wikstrom87,milner90,jensen97a}. The $AGENT$ color identifies the agents participating in the
interaction, and is used to construct the two compound color sets.

The \emph{INTER-STATE} color set is associated with agent places, and represents agents in the appropriate joint
interaction states. It is a record $\langle a_{1},a_{2}\rangle$, where $a_{1}$ and $a_{2}$ are $AGENT$ color
elements distinguishing the interacting agents. We apply the \emph{INTER-STATE} color set to model multiple
concurrent conversations using the same CP-net. The second color set is $MSG$, describing interaction
communicative acts and associated with message places. The $MSG$ color token is a record $\langle
a_{s},a_{r}\rangle$, where $a_{s}$ and $a_{r}$ correspond to the sender and the receiver agents of the
associated communicative act. In both cases, additional elements, such as conversation identification, may be
used. See Section~\ref{interaction-attributes} for additional details.

In Figure~\ref{fig-asynch-msg}-b, the $A_{1}B_{1}$ and the $A_{2}B_{2}$ places are associated with the
\emph{INTER-STATE} color set, while the $msg$ place is associated with the $MSG$ color set. The place color set
is written in italic capital letters next to the corresponding place. Furthermore, we use the $s$ and $r$
$AGENT$ color type variables to denote the net arc expressions. Thus, given that the output arc expression of
both the $A_{1}B_{1}$ and the $msg$ places is $\langle s,r\rangle$, the $s$ and $r$ elements of the agent place
token must correspond to the $s$ and $r$ elements of the message place token. Consequently, the net transition
occurs if and only if the agents of the message correspond to the interacting agents. The $A_{2}B_{2}$ place
input arc expression is $\langle r,s\rangle$ following the underlying intuition that $agent_{2}$ is going to
send the next interaction communicative act.

\begin{figure}[ht]
\begin{center}
\subfigure[AUML representation]{
\includegraphics[width=0.35\columnwidth,keepaspectratio]{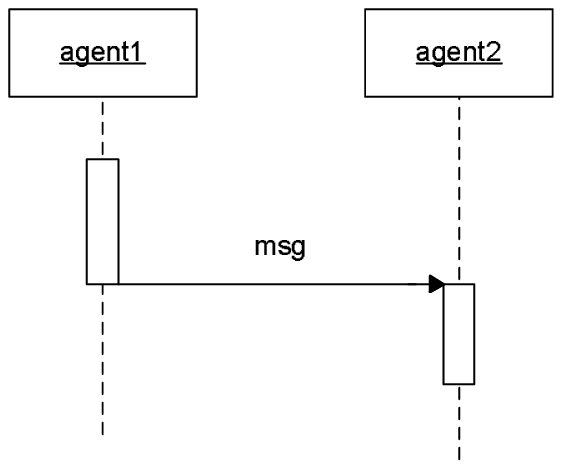}}
\subfigure[CP-net representation]{
\includegraphics[width=0.40\columnwidth,keepaspectratio]{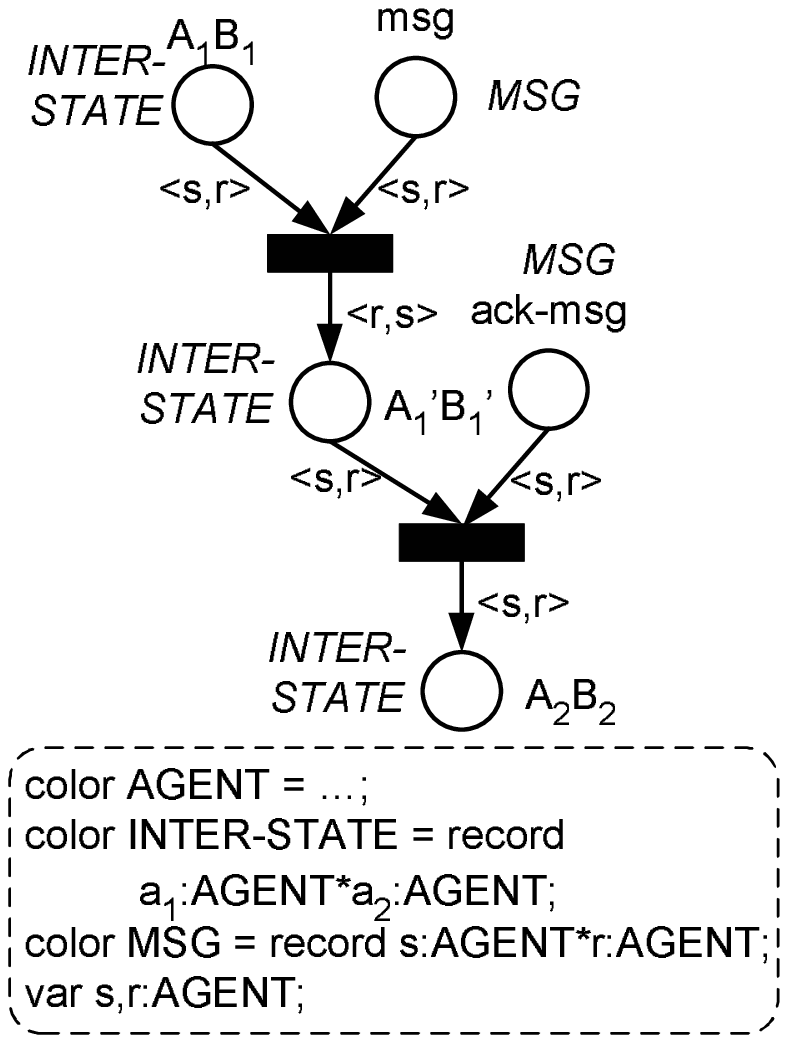}}
\caption {Synchronous message interaction.}
  \label{fig-synch-msg}
\vspace{-21pt}
  \end{center}
\end{figure}

Figure~\ref{fig-synch-msg}-a shows an AUML representation of another interaction building block, synchronous
message passing, denoted by the filled solid arrowhead. Here, the $msg$ communicative act is sent synchronously
from $agent_{1}$ to $agent_{2}$, meaning that an acknowledgement on $msg$ communicative act must always be
received by $agent_{1}$ before the interaction may proceed.

The corresponding CP-net representation is shown in Figure~\ref{fig-synch-msg}-b. The interaction starts in the
$A_{1}B_{1}$ place and terminates in the $A_{2}B_{2}$ place. The $A_{1}B_{1}$ place represents a joint
interaction state where $agent_{1}$ is ready to send the $msg$ communicative act to $agent_{2}$ ($A_{1}$) and
$agent_{2}$ is waiting to receive the corresponding message ($B_{1}$). The $A_{2}B_{2}$ place denotes a joint
interaction state, in which $agent_{1}$ has already sent the $msg$ communicative act to $agent_{2}$ ($A_{2}$)
and $agent_{2}$ has received it ($B_{2}$). However, since the CP-net diagram represents synchronous message
passing, the $msg$ communicative act transmission cannot cause the agents to transition directly from the
$A_{1}B_{1}$ place to the $A_{2}B_{2}$ place. We therefore define an intermediate $A_{1}'B_{1}'$ agent place.
This place represents a joint interaction state where $agent_{2}$ has received the $msg$ communicative act and
is ready to send an acknowledgement on it ($B_{1}$'), while $agent_{1}$ is waiting for that acknowledgement
($A_{1}'$). Taken together, the $msg$ communicative act causes the agents to transition from the $A_{1}B_{1}$
place to the $A_{1}'B_{1}'$ place, while the acknowledgement on the $msg$ message causes the agents to
transition from the $A_{1}'B_{1}'$ place to the $A_{2}B_{2}$ place.

Transitions in a typical multi-agent interaction protocols are composed of interaction building blocks, two of
which have been presented above.  Additional interaction building-blocks, which are fairly straightforward
\cite<or have appeared in previous work, e.g.,>{poutakidis02} are presented in
Appendix~\ref{additional-examples}. In the remainder of this section, we present two complex interactions
building blocks that are generally common in multi-agent interactions: XOR-decision and OR-parallel.

\begin{figure}[ht]
\begin{center}
\subfigure[AUML representation]{
\includegraphics[width=0.35\columnwidth,keepaspectratio]{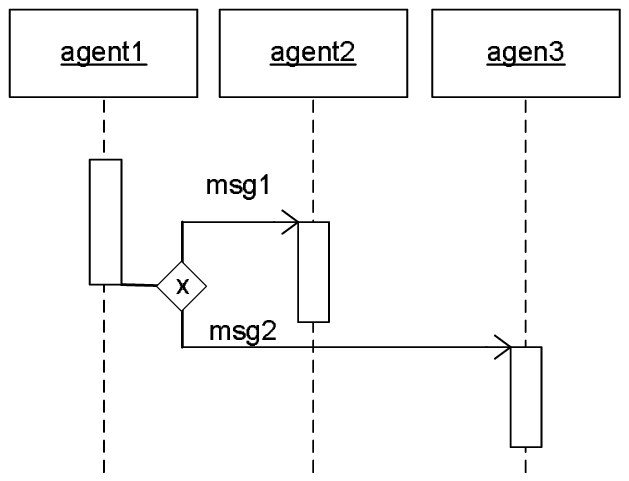}}
\subfigure[CP-net representation]{
\includegraphics[width=0.62\columnwidth,keepaspectratio]{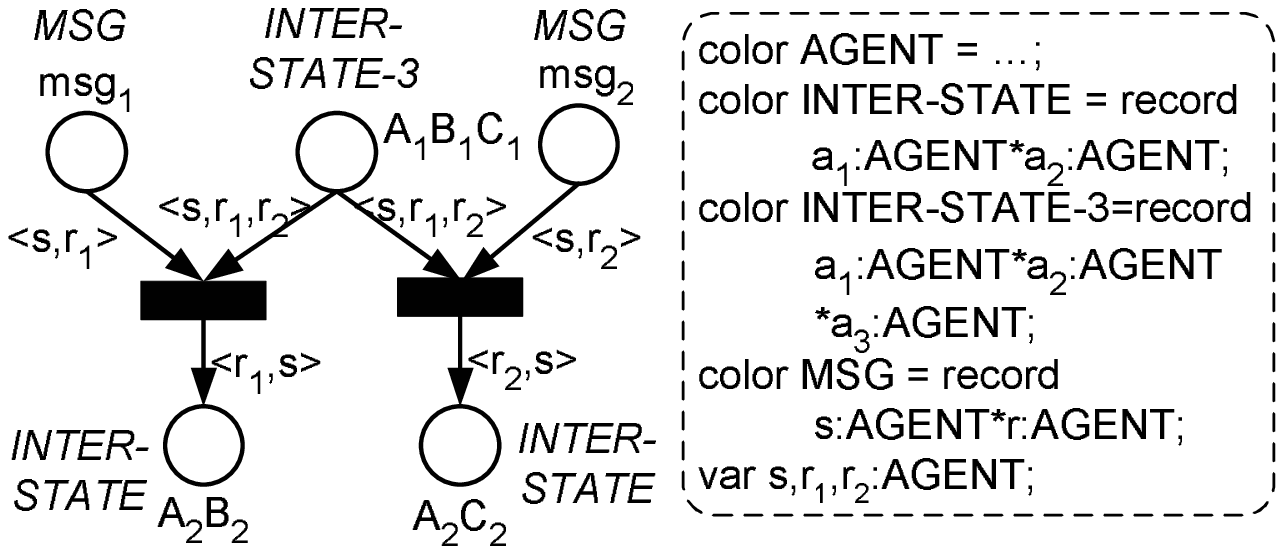}}
\caption {XOR-decision messages interaction.}
  \label{fig-xor-msg}
\vspace{-20pt}
  \end{center}
\end{figure}

We begin with the XOR-decision interaction. The AUML representation to this building block is shown in
Figure~\ref{fig-xor-msg}-a. The sender agent $agent_{1}$ can either send message $msg_{1}$ to $agent_{2}$ or
message $msg_{2}$ to $agent_{3}$, but it can not send both $msg_{1}$ and $msg_{2}$. The non-filled diamond with
an 'x' inside is the AUML notation for this constraint.

Figure~\ref{fig-xor-msg}-b shows the corresponding CP-net. Again, the $A$, $B$ and $C$ capital letters are used
to denote the interaction states of $agent_{1}$, $agent_{2}$ and $agent_{3}$, respectively. The interaction
starts from the $A_{1}B_{1}C_{1}$ place and terminates either in the $A_{2}B_{2}$ place or in the $A_{2}C_{2}$
place. The $A_{1}B_{1}C_{1}$ place represents a joint interaction state where $agent_{1}$ is ready to send
either the $msg_{1}$ communicative act to $agent_{2}$ or the $msg_{2}$ communicative act to $agent_{3}$
($A_{1}$); and $agent_{2}$ and $agent_{3}$ are waiting to receive the corresponding $msg_{1}$/$msg_{2}$ message
($B_{1}$/$C_{1}$). To represent the $A_{1}B_{1}C_{1}$ place color set, we extend the \emph{INTER-STATE} color
set to denote a joint interaction state of three interacting agents, i.e. using the \emph{INTER-STATE-3} color
set. The $msg_{1}$ communicative act causes the agents to transition to $A_{2}B_{2}$ place. The $A_{2}B_{2}$
place represents a joint interaction state where $agent_{1}$ has sent the $msg_{1}$ message ($A_{2}$), and
$agent_{2}$ has received it ($B_{2}$). Similarly, the $msg_{2}$ communicative act causes agents $agent_{1}$ and
$agent_{3}$ to transition to $A_{2}C_{2}$ place. Exclusiveness is achieved since the single agent token in
$A_{1}B_{1}C_{1}$ place can be used either for activating the $A_{1}B_{1}C_{1}\rightarrow A_{2}B_{2}$ transition
or for activating the $A_{1}B_{1}C_{1}\rightarrow A_{2}C_{2}$ transition, but not both.

\begin{figure}[ht]
\begin{center}
\subfigure[AUML representation]{
\includegraphics[width=0.35\columnwidth,keepaspectratio]{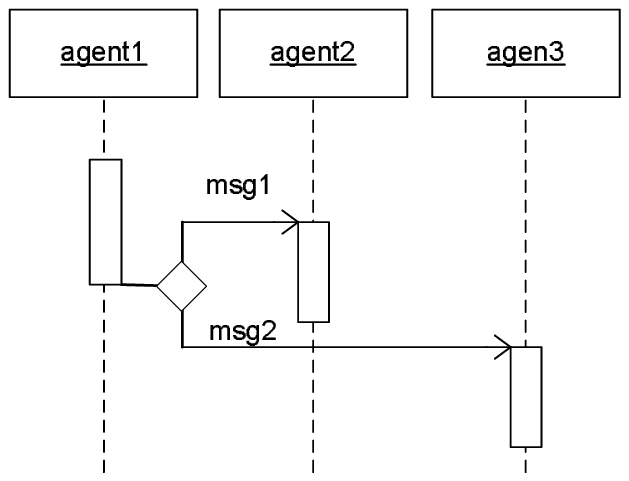}}
\subfigure[CP-net representation]{
\includegraphics[width=0.62\columnwidth,keepaspectratio]{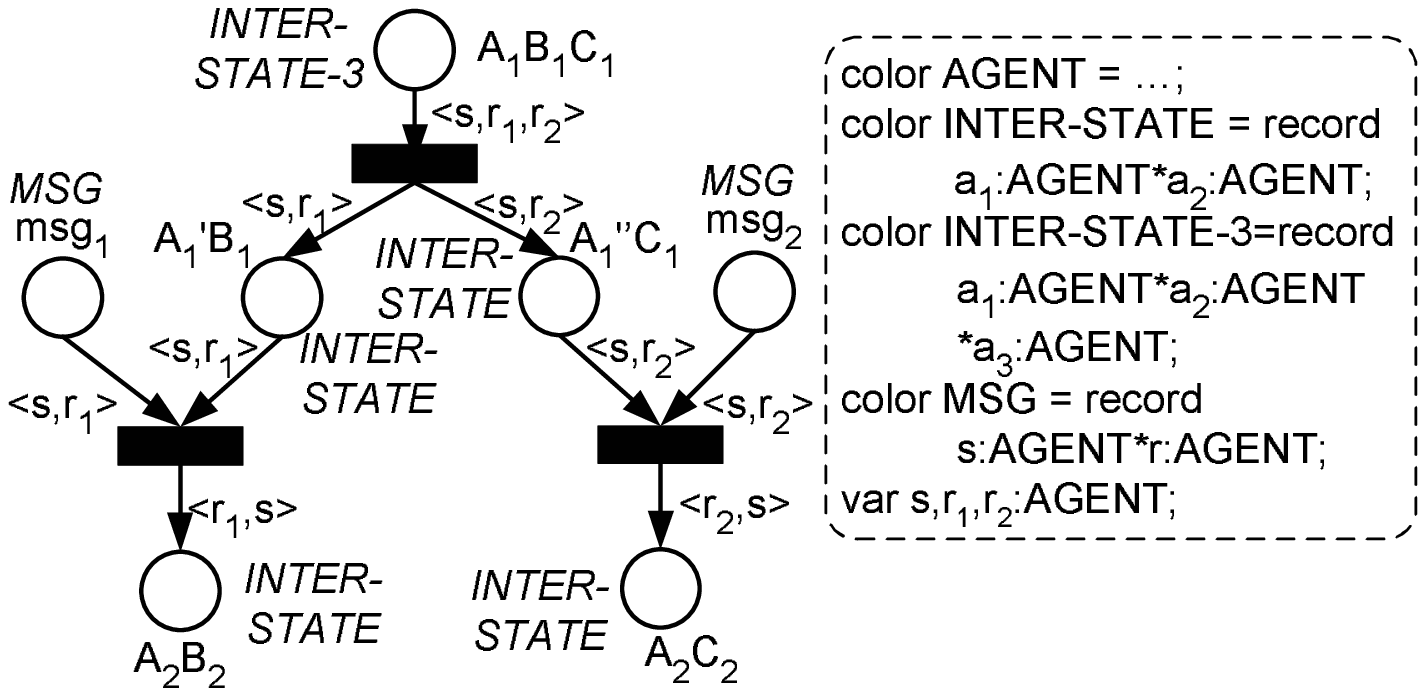}}
\caption {OR-parallel messages interaction.}
  \label{fig-or-parallel-msg}
\vspace{-15pt}
  \end{center}
\end{figure}

A similar complex interaction is the OR-parallel messages interaction. Its AUML representation is presented in
Figure~\ref{fig-or-parallel-msg}-a. The sender agent, $agent_{1}$, can send message $msg_{1}$ to $agent_{2}$ or
message $msg_{2}$ to $agent_{3}$, or both. The non-filled diamond is the AUML notation for this constraint.

Figure~\ref{fig-or-parallel-msg}-b shows the CP-net representation of the OR-parallel interaction. The
interaction starts from the $A_{1}B_{1}C_{1}$ place but it can be terminated in the $A_{2}B_{2}$ place, or in
the $A_{2}C_{2}$ place, or in both. To represent this inclusiveness of the interaction protocol, we define two
intermediate places, the $A_{1}'B_{1}$ place and the $A_{1}''C_{1}$ place. The $A_{1}'B_{1}$ place represents a
joint interaction state where $agent_{1}$ is ready to send the $msg_{1}$ communicative act to $agent_{2}$
($A_{1}'$) and $agent_{2}$ is waiting to receive the message ($B_{1}$). The $A_{1}''C_{1}$ place has similar
meaning, but with respect to $agent_{3}$. As normally done in Petri nets, the transition connecting the
$A_{1}B_{1}C_{1}$ place to the intermediate places duplicates any single token in $A_{1}B_{1}C_{1}$ place into
two tokens going into the $A_{1}'B_{1}$ and the $A_{1}''C_{1}$ places. Consequently, the two parts of the
OR-parallel interaction can be independently executed.

%-------------------------------------------------------------------------
\section{Representing Interaction Attributes}
\label{interaction-attributes}

We now extend our representation to allow additional interaction aspects, useful in describing multi-agent
conversation protocols. First, we show how to represent interaction message attributes, such as guards, sequence
expressions, cardinalities and content \cite{fipaa}. We then explore in depth the representation of multiple
concurrent conversations (on the same CP net).

\begin{figure}[ht]
\begin{center}
\subfigure[AUML representation]{
\includegraphics[width=0.30\columnwidth,keepaspectratio]{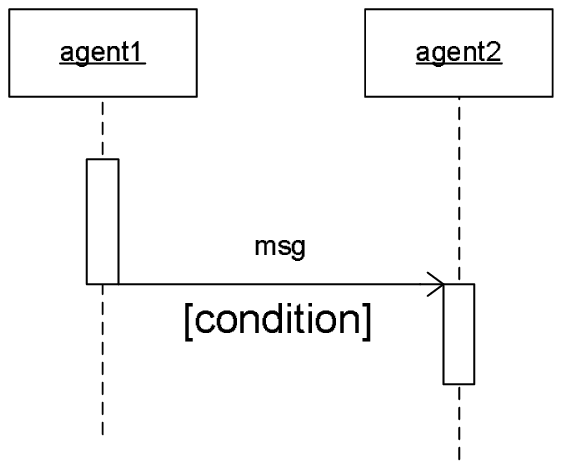}}
\hspace{0.005\columnwidth} \subfigure[CP-net representation]{
\includegraphics[width=0.65\columnwidth,keepaspectratio]{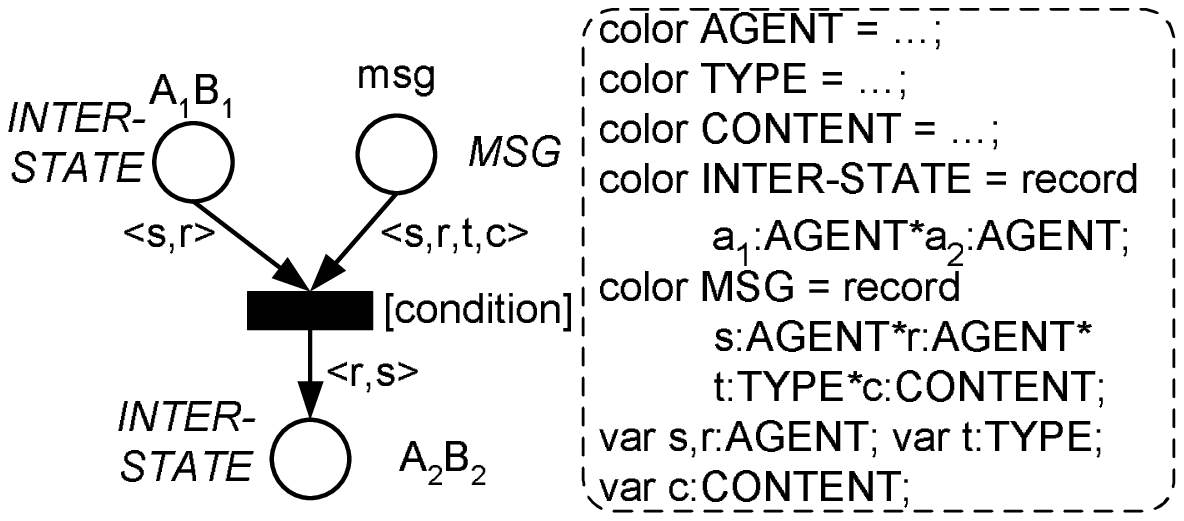}}
\caption {Message guard-condition}
  \label{fig-condition-msg}
\vspace{-15pt}
  \end{center}
\end{figure}

Figure~\ref{fig-condition-msg}-a shows a simple agent interaction using an AUML protocol diagram. This
interaction is similar to the one presented in Figure~\ref{fig-asynch-msg}-a in the previous section. However,
Figure~\ref{fig-condition-msg}-a uses an AUML message guard-condition--marked as $[condition]$--that has the
following semantics: the communicative act is sent from $agent_{1}$ to $agent_{2}$ if and only if the
$condition$ is true.

The guard-condition implementation in our Petri net representation uses transition guards
(Figure~\ref{fig-condition-msg}-b), a native feature for CP nets. The AUML guard condition is mapped directly to
the CP-net transition guard. The CP-net transition guard is indicated on the net inscription next to the
corresponding transition using square brackets. The transition guard guarantees that the transition is enabled
if and only if the transition guard is true.

In Figure~\ref{fig-condition-msg}-b, we also extend the color of tokens to include information about the
communicative act being used and its content. We extend the $MSG$ color set definition to a record $\langle
s,r,t,c\rangle$, where the $s$ and $r$ elements has the same interpretation as in previous section (sender and
receiver), and the $t$ and $c$ elements define the message type and content, respectively. The $t$ element is of
a new color $TYPE$, which determines communicative act types. The $c$ element is of a new color $CONTENT$, which
represents communicative act content and argument list (e.g. reply-to, reply-by and etc).

The addition of new elements also allows for additional potential uses. For instance, to facilitate
representation of multiple concurrent conversations between the same agents ($s$ and $r$), it is possible to add
a conversation identification field to both the $MSG$ and \emph{INTER-STATE} colors. For simplicity, we refrain
from doing so in the examples in this paper.

Two additional AUML communicative act attributes that can be modelled in the CP representation are message
sequence-expression and message cardinality. The sequence-expressions denote a constraint on the message sent
from sender agent. There are a number of sequence-expressions defined by FIPA conversation standards
\cite{fipaa}: $m$ denotes that the message is sent exactly $m$ times; $n..m$ denotes that the message is sent
anywhere from $n$ up to $m$ times; ${*}$ denotes that the message is sent an arbitrary number of times. An
additional important sequence expression is $broadcast$, i.e. message is sent to all other agents.

\begin{figure}[ht]
\begin{center}
\subfigure[AUML representation]{
\includegraphics[width=0.30\columnwidth,keepaspectratio]{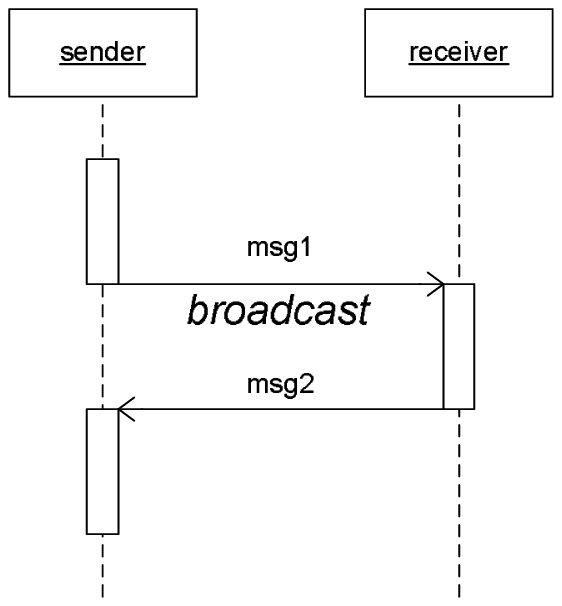}}
 \subfigure[CP-net representation]{
\includegraphics[width=0.67\columnwidth,keepaspectratio]{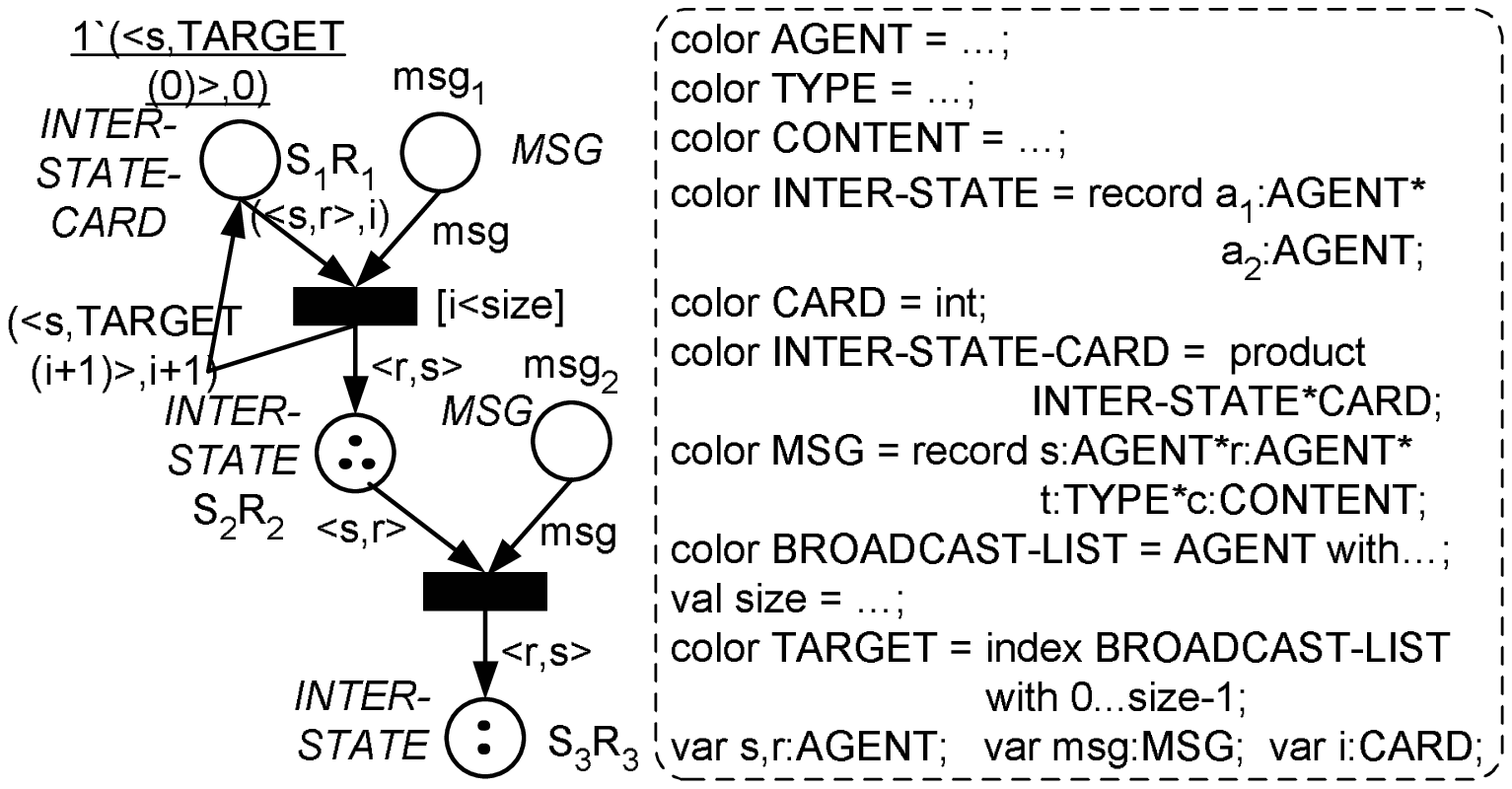}}
\caption {Broadcast sequence expression.}
  \label{fig-broadcast-msg}
\vspace{-20pt}
  \end{center}
\end{figure}

We now explain the representation of sequence-expressions in CP-nets, using broadcast as an example
(Figure~\ref{fig-broadcast-msg}-b). Other sequence expressions are easily derived from this example. We define
an \emph{INTER-STATE-CARD} color set. This color set is a tuple $(\langle a_{1},a_{2}\rangle,i)$ consisting of
two elements. The first tuple element is an \emph{INTER-STATE} color element, which denotes the interacting
agents as previously defined. The second tuple element is an integer that counts the number of messages already
sent by an agent, i.e. the message cardinality. This element is initially assigned to 0. The
\emph{INTER-STATE-CARD} color set is applied to the $S_{1}R_{1}$ place, where the $S$ and $R$ capital letters
are used to denote the $sender$ and the $receiver$ individual interaction states respectively and the
$S_{1}R_{1}$ indicates the initial joint interaction state of the interacting agents. The two additional colors,
used in Figure~\ref{fig-broadcast-msg}-b, are the \emph{BROADCAST-LIST} and the $TARGET$ colors. The
\emph{BROADCAST-LIST} color defines the sender broadcast list of the designated receivers, assuming that the
$sender$ must have such a list to carry out its role. The $TARGET$ color defines indexes into this broadcast
list.

According to the broadcast sequence-expression semantics, the $sender$ agent sends the same $msg_{1}$
communicative act to all the $receivers$ on the $broadcast\ list$. The CP-net introduced in
Figure~\ref{fig-broadcast-msg}-b models this behavior.\footnote{ We implement broadcast as an iterative
procedure sending the corresponding communicative act separately to all designated recipients.} The interaction
starts from the $S_{1}R_{1}$ place, representing the joint interaction state where $sender$ is ready to send the
$msg_{1}$ communicative act to $receiver$ ($S_{1}$) and $receiver$ is waiting to receive the corresponding
$msg_{1}$ message ($R_{1}$). The $S_{1}R_{1}$ place initial marking is a single token, set by the initialization
expression (underlined, next to the corresponding place). The initialization expression $1`(\langle
s,TARGET(0)\rangle,0)$--given in standard CPN ML notation--determines that the $S_{1}R_{1}$ place's initial
marking is a multi-set containing a single token $(\langle s,TARGET(0)\rangle,0)$. Thus, the first designated
$receiver$ is assigned to be the agent with index $0$ on the broadcast list, and the message cardinality counter
is initiated to $0$.

The $msg_{1}$ message place initially contains multiple tokens. Each of these tokens represents the $msg_{1}$
communicative act addressed to a different designated $receiver$ on the $broadcast \ list$. In
Figure~\ref{fig-broadcast-msg}-b, the initialization expression, corresponding to the $msg_{1}$ message place,
has been omitted. The $S_{1}R_{1}$ place token and the appropriate $msg_{1}$ place token together enable the
corresponding transition. Consequently, the transition may fire and thus the $msg_{1}$ communicative act
transmission is simulated.

The $msg_{1}$ communicative act is sent incrementally to every designated $receiver$ on the $broadcast\ list$.
The incoming arc expression $(\langle s,r\rangle,i)$ is incremented by the transition to the outgoing $(\langle
s,TARGET(i+1)\rangle,i+1)$ arc expression, causing the $receiver$ agent with index $i+1$ on the $broadcast\
list$ to be selected. The transition guard constraint $i<size$, i.e. $i<|broadcast\ list|$, ensures that the
$msg_{1}$ message is sent no more than $|broadcast\ list|$ times. The $msg_{1}$ communicative act causes the
agents to transition to the $S_{2}R_{2}$ place. This place represents a joint interaction state in which sender
has already sent the $msg_{1}$ communicative act to $receiver$ and is now waiting to receive the $msg_{2}$
message ($S_{2}$) and $receiver$ has received the $msg_{1}$ message and is ready to send the $msg_{2}$
communicative act to sender ($R_{2}$). Finally, the $msg_{2}$ message causes the agents to transition to the
$S_{3}R_{3}$ place. The $S_{3}R_{3}$ place denotes a joint interaction state where sender has received the
$msg_{2}$ communicative act from $receiver$ and terminated ($S_{3}$), while $receiver$ has already sent the
$msg_{2}$ message to $sender$ and terminated as well ($R_{3}$).

We use Figure~\ref{fig-broadcast-msg}-b to demonstrate the use of token color to represent multiple concurrent
conversations using the same CP-net. For instance, let us assume that the $sender$ agent is called $agent_{1}$
and its $broadcast\ list$ contains the following agents: $agent_{2}$, $agent_{3}$, $agent_{4}$, $agent_{5}$ and
$agent_{6}$. We will also assume that the $agent_{1}$ has already sent the $msg_{1}$ communicative act to all
agents on the $broadcast\ list$. However, it has only received the $msg_{2}$ reply message from $agent_{3}$ and
$agent_{6}$. Thus, the CP-net current marking for the complete interaction protocol is described as follows: the
$S_{2}R_{2}$ place is marked by {$\langle agent_{2}, agent_{1}\rangle$, $\langle agent_{4},agent_{1}\rangle$,
$\langle agent_{5}, agent_{1}\rangle$}, while the $S_{3}R_{3}$ place contains the tokens {$\langle
agent_{1},agent_{3}\rangle$ and $\langle agent_{1},agent_{6}\rangle$}.

\begin{figure} [ht]
  \begin{center}
\includegraphics[width=0.37\columnwidth,keepaspectratio]{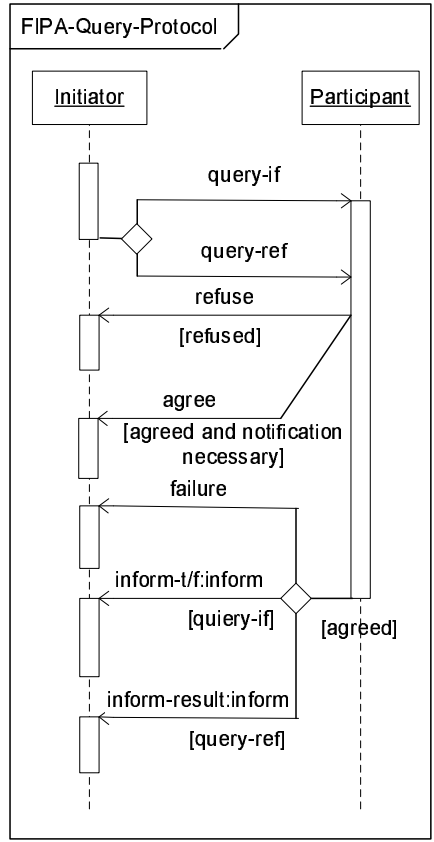}\\
  \caption {FIPA Query Interaction Protocol - AUML
representation.}\label{fig-query-auml} %\vspace{-25pt}
  \end{center}
\end{figure}

\textbf{An Example}. We now construct a CP-net representation of the \emph{FIPA Query Interaction Protocol}
\cite{fipab}, shown in AUML form in Figure~\ref{fig-query-auml}, to demonstrate how the building blocks
presented in Sections~\ref{simple-and-complex-building-blocks} and \ref{interaction-attributes} can be put
together. In this interaction protocol, the $Initiator$ requests the $Participant$ to perform an inform action
using one of two query communicative acts, \emph{query-if} or \emph{query-ref}. The $Participant$ processes the
query and makes a decision whether to $accept$ or $refuse$ the query request. The $Initiator$ may request the
$Participant$ to respond with either an $accept$ or $refuse$ message, and for simplicity, we will assume that
this is always the case. In case the $query$ request has been accepted, the $Participant$ informs the
$Initiator$ on the query results. If the $Participant$ fails, then it communicates a $failure$. In a successful
response, the $Participant$ replies with one of two versions of inform (\emph{inform-t/f} or
\emph{inform-result}) depending on the type of initial query request.

\begin{figure}[ht]
  \begin{center}

\includegraphics[width=0.87\columnwidth,keepaspectratio]{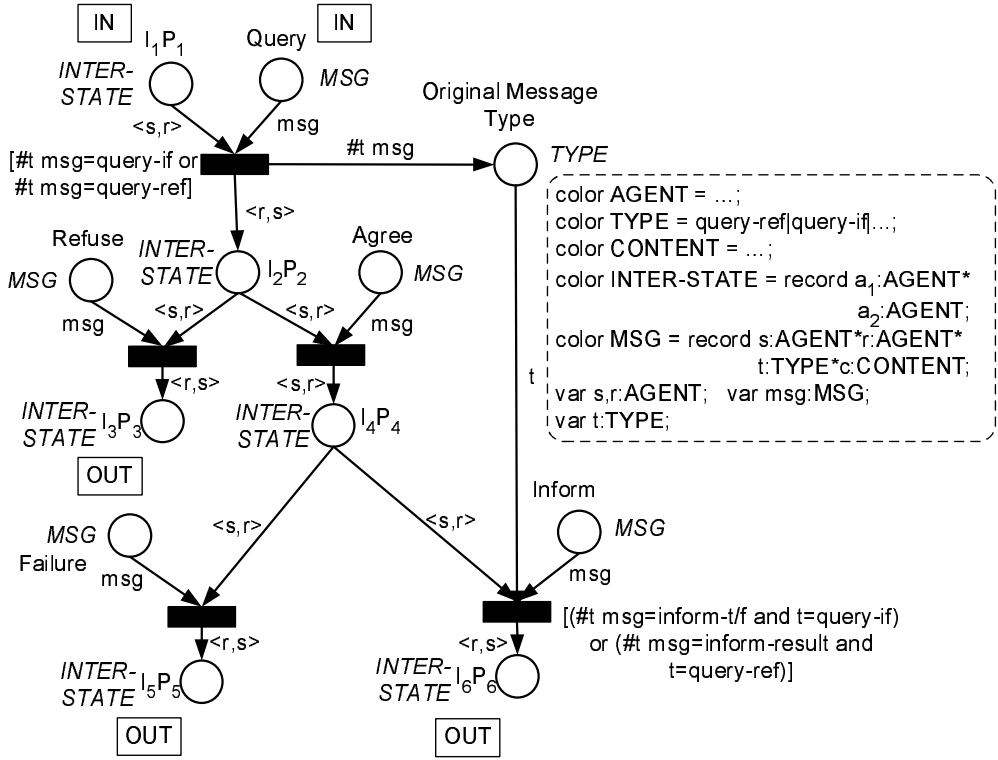}\\
  \caption {FIPA Query Interaction Protocol - CP-net
representation.}\label{fig-query-cp} \vspace{-15pt}
  \end{center}
\end{figure}

The CP-net representation of the \emph{FIPA Query Interaction Protocol} is presented in
Figure~\ref{fig-query-cp}. The interaction starts in the $I_{1}P_{1}$ place (we use the $I$ and the $P$ capital
letters to denote the $Initiator$ and the $Participant$ roles). The $I_{1}P_{1}$ place represents a joint
interaction state where (i) the $Initiator$ agent is ready to send either the \emph{query-if} communicative act,
or the \emph{query-ref} message, to $Participant$ ($I_{1}$); and (ii) $Participant$ is waiting to receive the
corresponding message ($P_{1}$). The $Initiator$ can send either a \emph{query-if} or a \emph{query-ref}
communicative act. We assume that these acts belong to the same class, the $query$ communicative act class.
Thus, we implement both messages using a single $Query$ message place, and check the message type using the
following transition guard: $[\#t\ msg=\mathrm{\emph{query-if}}\ \mathrm{\ or\ }\ \# t\
msg=\mathrm{\emph{query-ref}}]$. The query communicative act causes the interacting agents to transition to the
$I_{2}P_{2}$ place. This place represents a joint interaction state in which $Initiator$ has sent the $query$
communicative act and is waiting to receive a response message ($I_{2}$), and $Participant$ has received the
$query$ communicative act and deciding whether to send an $agree$ or a $refuse$ response message to $Initiator$
($P_{2}$). The $refuse$ communicative act causes the agents to transition to $I_{3}P_{3}$ place, while the
$agree$ message causes the agents to transition to $I_{4}P_{4}$ place.

The $Participant$ decision on whether to send an $agree$ or a $refuse$ communicative act is represented using
the XOR-decision building block introduced earlier (Figure~\ref{fig-xor-msg}-b). The $I_{3}P_{3}$ place
represents a joint interaction state where $Initiator$ has received a $refuse$ communicative act and terminated
($I_{3}$) and $Participant$ has sent a $refuse$ message and terminated as well ($P_{3}$). The $I_{4}P_{4}$ place
represents a joint interaction state in which $Initiator$ has received an $agree$ communicative act and is now
waiting for further response from $Participant$ ($I_{4}$) and $Participant$ has sent an $agree$ message and is
now deciding which response to send to $Initiator$ ($P_{4}$). At this point, the $Participant$ agent may send
one of the following communicative acts: \emph{inform-t/f}, \emph{inform-result} and $failure$. The choice is
represented using another XOR-decision building block, where the \emph{inform-t/f} and \emph{inform-result}
communicative acts are represented using a single $Inform$ message place. The $failure$ communicative act causes
a transition to the $I_{5}P_{5}$ place, while the $inform$ message causes a transition to the $I_{6}P_{6}$
place. The $I_{5}P_{5}$ place represents a joint interaction state where $Participant$ has sent a $failure$
message and terminated ($P_{5}$), while $Initiator$ has received a $failure$ and terminated ($I_{5}$). The
$I_{6}P_{6}$ place represents a joint interaction state in which $Participant$ has sent an $inform$ message and
terminated ($P_{6}$), while $Initiator$ has received an $inform$ and terminated ($I_{6}$).

The implementation of the [\emph{query-if}] and the [\emph{query-ref}] message guard conditions requires a
detailed discussion. These are not implemented in a usual manner in view of the fact that they depend on the
original request communicative act. Thus, we create a special intermediate place that contains the original
message type marked "$Original\ Message\ Type$" in the figure. In case an $inform$ communicative act is sent,
the transition guard verifies that the $inform$ message is appropriate to the original $query$ type. Thus, an
\emph{inform-t/f} communicative act can be sent only if the original query type has been \emph{query-if} and an
\emph{inform-result} message can be sent only if the original query type has been \emph{query-ref}.

%-------------------------------------------------------------------------
\section{Representing Nested \& Interleaved Interactions}
\label{nested-and-interleaved-interactions}

In this section, we extend the CP-net representation of previous sections to model nested and interleaved
interaction protocols. We focus here on nested interaction protocols. Nevertheless, the discussion can also be
addressed to interleaved interaction protocols in a similar fashion.

FIPA conversation standards \cite{fipaa} emphasize the importance of nested and interleaved protocols in
modelling complex interactions. First, this allows re-use of interaction protocols in different nested
interactions. Second, nesting increases the readability of interaction protocols.

\begin{figure}[ht]
\begin{center}
\subfigure[Nested protocol]{
\includegraphics[width=0.37\columnwidth,keepaspectratio]{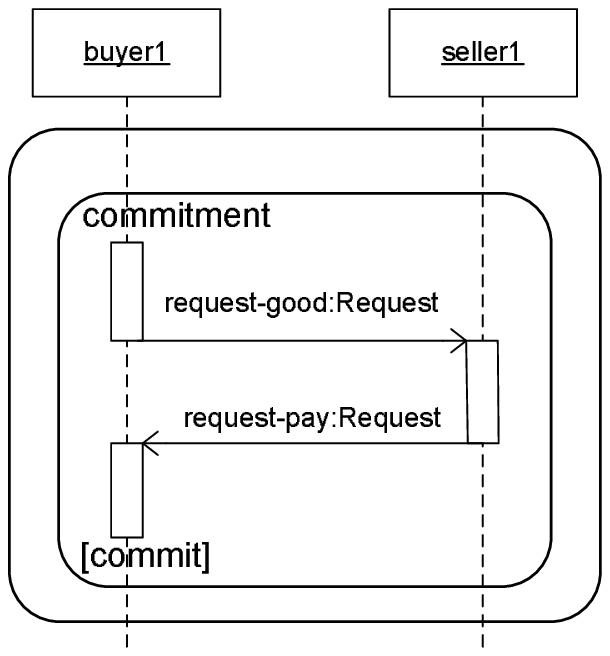}}
\hspace{0.002\columnwidth} \subfigure[Interleave protocol]{
\includegraphics[width=0.57\columnwidth,keepaspectratio]{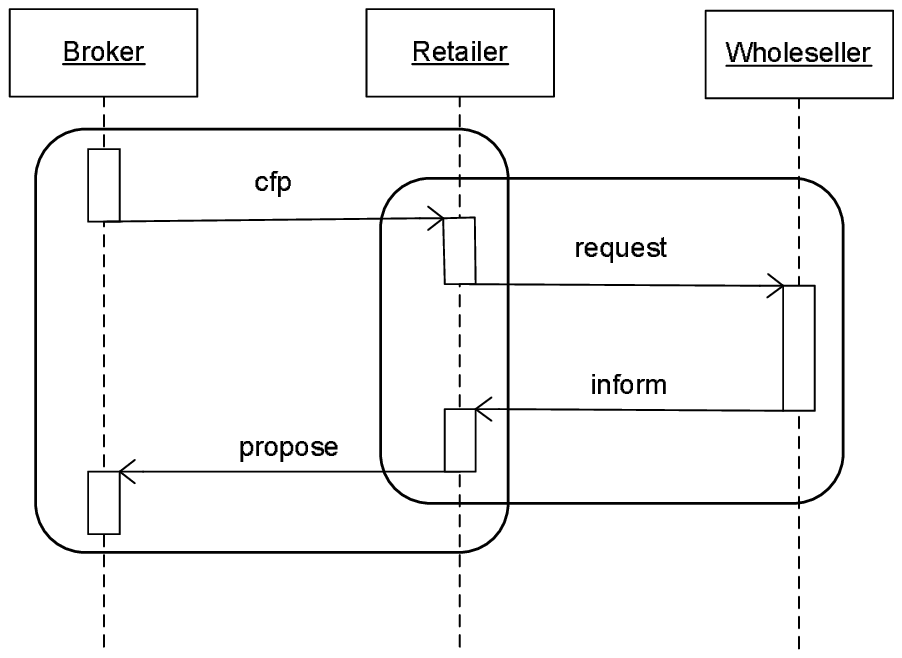}}
\caption {AUML nested and interleaved protocols examples.}
  \label{fig-nested-auml}
\vspace{-15pt}
  \end{center}
\end{figure}

The AUML notation annotates nested and interleaved protocols as round corner rectangles \cite{odell01b,fipaa}.
Figure~\ref{fig-nested-auml}-a shows an example of a nested protocol\footnote{Figure~\ref{fig-nested-auml}-a
appears in FIPA conversation standards \cite{fipaa}. Nonetheless, note that the \emph{request-good} and the
\emph{request-pay} communicative acts are not part of the FIPA-ACL standards.}, while
Figure~\ref{fig-nested-auml}-b illustrates an interleaved protocol. Nested protocols have one or more
compartments. The first compartment is the name compartment. The name compartment holds the (optional) name of
the nested protocol. The nested protocol name is written in the upper left-hand corner of the rectangle, i.e.
$commitment$ in Figure~\ref{fig-nested-auml}-a. The second compartment, the guard compartment, holds the
(optional) nested protocol guard. The guard compartment is written in the lower left-hand corner of the
rectangle, e.g. $[commit]$ in Figure~\ref{fig-nested-auml}-a. Nested protocols without guards are equivalent to
nested protocols with the $[true]$ guard.

\begin{figure}[ht]
  \begin{center}
\includegraphics[width=0.9\columnwidth,keepaspectratio]{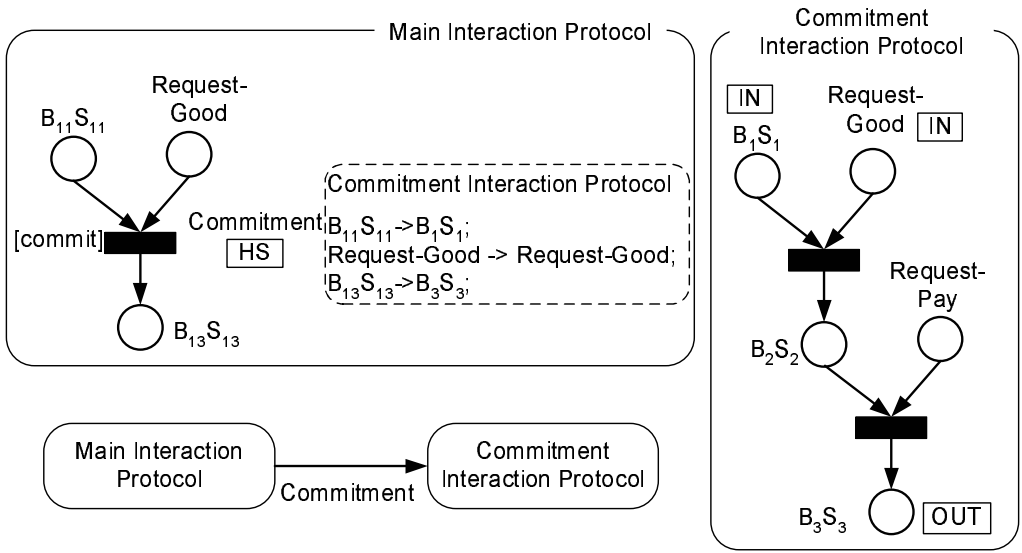}\\
  \caption {Nested protocol implementation using hierarchical
CP-nets.}\label{fig-nested-cp} \vspace{-15pt}
  \end{center}
\end{figure}

Figure~\ref{fig-nested-cp} describes the implementation of the nested interaction protocol presented in
Figure~\ref{fig-nested-auml}-a by extending the CP-net representation to using hierarchies, relying on standard
CP-net methods (see Appendix~\ref{petri-nets-introduction}). The hierarchical CP-net representation contains
three elements: a \emph{superpage}, a \emph{subpage} and a \emph{page hierarchy} graph. The CP-net superpage
represents the main interaction protocol containing a nested interaction, while the CP-net subpage models the
corresponding nested interaction protocol, i.e. the \emph{Commitment Interaction Protocol}. The page hierarchy
graph describes how the superpage is decomposed into subpages.

Let us consider in detail the process of modelling the nested interaction protocol in
Figure~\ref{fig-nested-auml}-a using a hierarchical CP-net, resulting in the net described in
Figure~\ref{fig-nested-cp}. First, we identify the starting and ending points of the nested interaction
protocol. The starting point of the nested interaction protocol is where $Buyer_{1}$ sends a \emph{Request-Good}
communicative act to $Seller_{1}$. The ending point is where $Buyer_{1}$ receives a \emph{Request-Pay}
communicative act from $Seller_{1}$. We model these nested protocol end-points as CP-net \emph{socket nodes} on
the superpage, i.e. $Main\ Interaction\ Protocol$: $B_{11}S_{11}$ and \emph{Request-Good} are input socket nodes
and $B_{13}S_{13}$ is an output socket node.

The nested interaction protocol, the $Commitment\ Interaction\ Protocol$, is represented using a separate
CP-net, following the principles outlined in Sections~\ref{simple-and-complex-building-blocks} and
\ref{interaction-attributes}. This net is a subpage of the main interaction protocol superpage. The nested
interaction protocol starting and ending points on the subpage correspond to the net \emph{port nodes}. The
$B_{1}S_{1}$ and \emph{Request-Good} places are the subpage input port nodes, while the $B_{3}S_{3}$ place is an
output port node. These nodes are tagged with the IN/OUT \emph{port type tags} correspondingly.

Then, a \emph{substitution transition}, which is denoted using HS (Hierarchy and Substitution), connects the
corresponding socket places on the superpage. The substitution transition conceals the nested interaction
protocol implementation from the net superpage, i.e. the $Main\ Interaction\ Protocol$. The nested protocol name
and guard compartments are mapped directly to the substitution transition name and guard respectively.
Consequently, in Figure~\ref{fig-nested-cp} we define the substitution transition name as $Commitment$ and the
substitution guard is determined to be $[commit]$.

The superpage and subpage interface is provided using the hierarchy inscription. The hierarchy inscription is
indicated using the dashed box next to the substitution transition. The first line in the hierarchy inscription
determines the subpage identity, i.e. the $Commitment\ Interaction\ Protocol$ in our example. Moreover, it
indicates that the substitution transition replaces the corresponding subpage detailed implementation on the
superpage. The remaining hierarchy inscription lines introduce the superpage and subpage port assignment. The
port assignment relates a socket node on the superpage with a port node on the subpage. The substitution
transition input socket nodes are related to the IN-tagged port nodes. Analogously, the substitution transition
output socket nodes correspond to the OUT-tagged port nodes. Therefore, the port assignment in
Figure~\ref{fig-nested-cp} assigns the net socket and port nodes in the following fashion: $B_{11}S_{11}$ to
$B_{1}S_{1}$, \emph{Request-Good} to \emph{Request-Good} and $B_{13}S_{13}$ to $B_{3}S_{3}$.

Finally, the page hierarchy graph describes the decomposition hierarchy (nesting) of the different protocols
(pages). The CP-net pages, the $Main\ Interaction\ Protocol$ and the $Commitment\ Interaction\ Protocol$,
correspond to the page hierarchy graph nodes (Figure~\ref{fig-nested-cp}). The arc inscription indicates the
substitution transition, i.e. $Commitment$.

%-------------------------------------------------------------------------
\section{Representing Temporal Aspects of Interactions}
\label{temporal-aspects}

Two temporal interaction aspects are specified by FIPA \cite{fipaa}. In this section, we show how \emph{timed}
CP-nets (see also Appendix~\ref{petri-nets-introduction}) can be applied for modelling agent interactions that
involve temporal aspects, such as interaction duration, deadlines for message exchange, etc.

A first aspect, \emph{duration}, is the interaction activity time period. Two periods can be distinguished:
\emph{transmission time} and \emph{response time}. The transmission time indicates the time interval during
which a communicative act, is sent by one agent and received by the designated receiver agent. The response time
period denotes the time interval in which the corresponding receiver agent is performing some task as a response
to the incoming communicative act.

The second temporal aspect is \emph{deadlines}. Deadlines denote the time limit by which a communicative act
must be sent. Otherwise, the corresponding communicative act is considered to be invalid. These issues have not
been addressed in previous investigations related to agent interactions modelling using Petri
nets.\footnote{\citeA{cost99b,cost00} mention deadlines without presenting any implementation details.}

\begin{figure} [ht]
\begin{center}
\subfigure[AUML representation]{
\includegraphics[width=0.30\columnwidth,keepaspectratio]{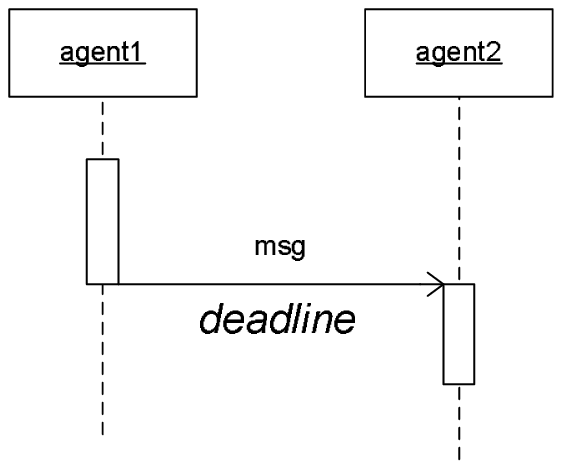}}
\hspace{0.005\columnwidth} \subfigure[CP-net representation]{
\includegraphics[width=0.65\columnwidth,keepaspectratio]{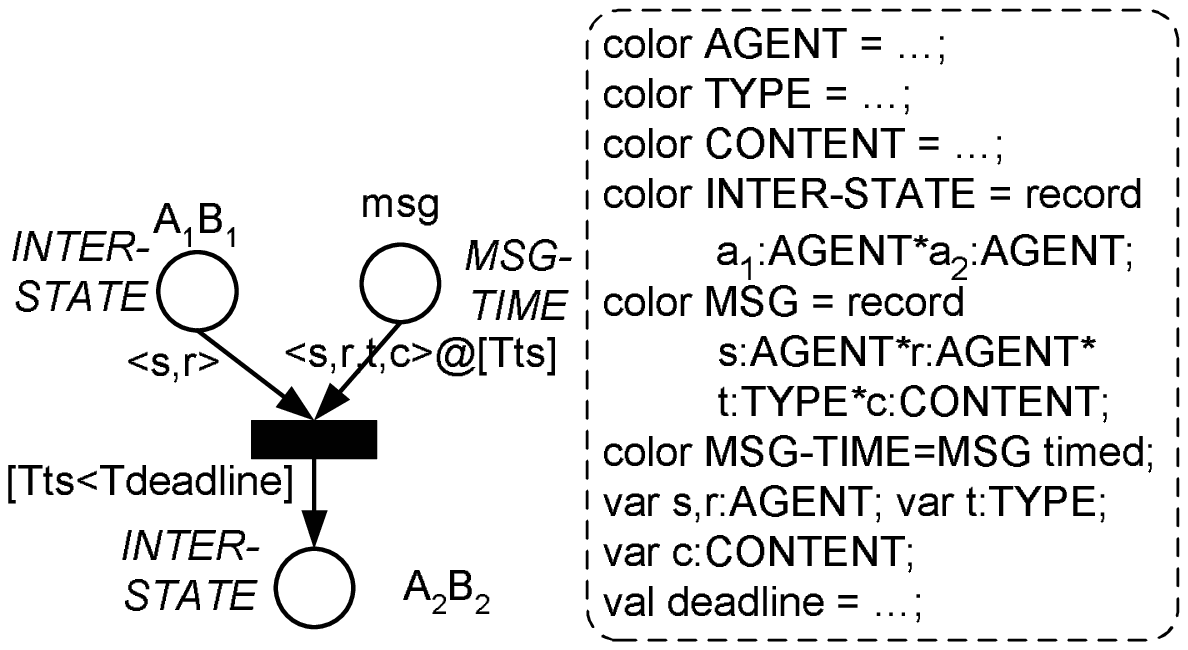}}
\caption {Deadline sequence expression.}
  \label{fig-deadline-msg}
\vspace{-20pt}
  \end{center}
\end{figure}

We propose to utilize timed CP-nets techniques to represent these temporal aspects of agent interactions. In
doing so, we assume a global clock.\footnote{Implementing it, we can use the private clock of an overhearing
agent as the global clock for our Petri net representation. Thus, the time stamp of the message is the
overhearer's time when the corresponding message was overheard.} We begin with deadlines.
Figure~\ref{fig-deadline-msg}-a introduces the AUML representation of message deadlines. The $deadline$ keyword
is a variation of the communicative act sequence expressions described in Section~\ref{interaction-attributes}.
It sets a time constraint on the start of the transmission of the associated communicative act. In
Figure~\ref{fig-deadline-msg}-a, $agent_{1}$ must send the $msg$ communicative act to $agent_{2}$ before the
defined $deadline$. Once the $deadline$ expires, the $msg$ communicative act is considered to be invalid.

Figure~\ref{fig-deadline-msg}-b shows a timed CP-net implementation of the deadline sequence expression. The
timed CP-net in Figure~\ref{fig-deadline-msg}-b defines an additional \emph{MSG-TIME} color set associated with
the net message places. The \emph{MSG-TIME} color set extends the $MSG$ color set, described in
Section~\ref{interaction-attributes}, by adding a time stamp attribute to the message token. Thus, the
communicative act token is a record $\langle s,r,t,c\rangle@[Tts]$. The $@[..]$ expression denotes the
corresponding token time stamp, whereas the token time value is indicated starting with a capital 'T'.
Accordingly, the described message token has a $ts$ time stamp. The communicative act time limit is defined
using the val $deadline$ parameter. Therefore, the deadline sequence expression semantics is simulated using the
following transition guard: $[Tts<Tdeadline]$. This transition guard, comparing the $msg$ time stamp against the
$deadline$ parameter, guarantees that an expired $msg$ communicative act can not be received.

\begin{figure}[ht]
\begin{center}
\subfigure[AUML representation]{
\includegraphics[width=0.30\columnwidth,keepaspectratio]{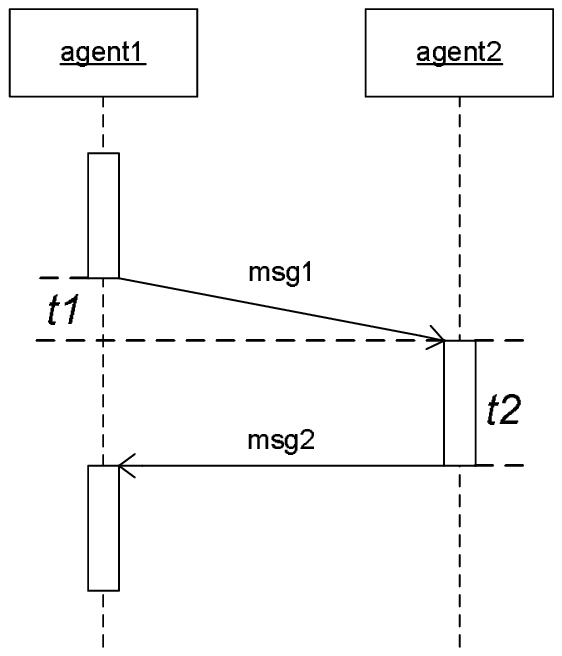}}
\hspace{0.005\columnwidth} \subfigure[CP-net representation]{
\includegraphics[width=0.65\columnwidth,keepaspectratio]{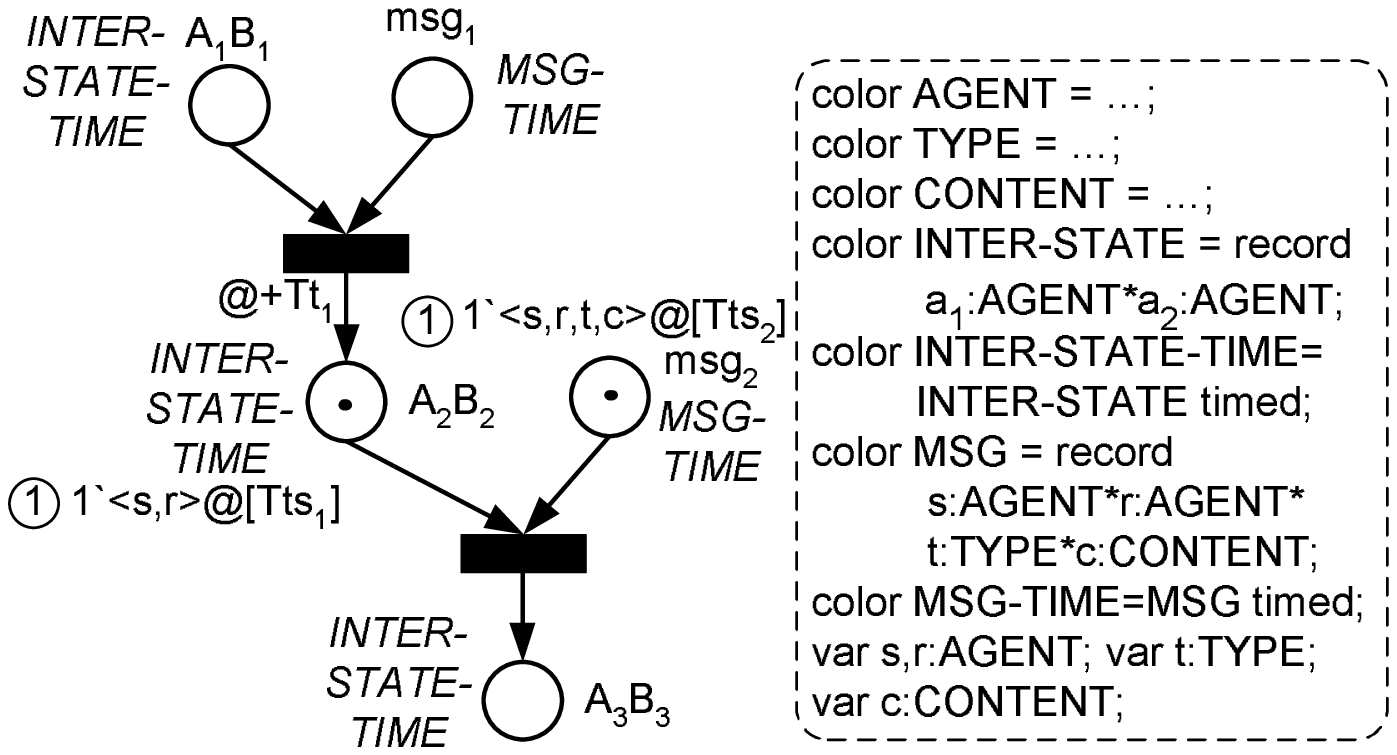}}
\caption {Interaction duration.}
  \label{fig-duration-msg}
\vspace{-20pt}
  \end{center}
\end{figure}

We now turn to representing interaction duration. The AUML representation is shown in
Figure~\ref{fig-duration-msg}-a. The AUML time intensive message notation is used to denote the communicative
act transmission time. As a rule communicative act arrows are illustrated horizontally. This indicates that the
message transmission time can be neglected. However, in case the message transmission time is significant, the
communicative act is drawn slanted downwards. The vertical distance, between the arrowhead and the arrow tail,
denotes the message transmission time. Thus, the communicative act $msg_{1}$, sent from $agent_{1}$ to
$agent_{2}$, has a $t_{1}$ transmission time.

The response time in Figure~\ref{fig-duration-msg}-a is indicated through the interaction thread length. The
incoming $msg_{1}$ communicative act causes $agent_{2}$ to perform some task before sending a response $msg_{2}$
message. The corresponding interaction thread duration is denoted through the $t_{2}$ time period. Thus, this
time period specifies the $agent_{2}$ response time to the incoming $msg_{1}$ communicative act.

The CP-net implementation to the interaction duration time periods is shown in Figure~\ref{fig-duration-msg}-b.
The communicative act transmission time is illustrated using the timed CP-nets @+ operator. The net transitions
simulate the communicative act transmission between agents. Therefore, representing a transmission time of
$t_{1}$, the CP-net transition adds a $t_{1}$ time period to the incoming message token time stamp. Accordingly,
the transition $@+Tt_{1}$ output arc expression denotes a $t_{1}$ delay to the time stamp of the outgoing token.
Thus, the corresponding transition takes $t_{1}$ time units and consequently so does the $msg_{1}$ communicative
act transmission time.

In contrast to communicative act transmission time, the agent interaction response time is represented
implicitly. Previously, we have defined a \emph{MSG-TIME} color set that indicates message token time stamps.
Analogously, in Figure~\ref{fig-duration-msg}-b we introduce an additional \emph{INTER-STATE-TIME} color set.
This color set is associated with the net agent places and it presents the possibility to attach time stamps to
agent tokens as well. Now, let us assume that $A_{2}B_{2}$ and $msg_{2}$ places contain a single token each. The
circled '1' next to the corresponding place, together with the multi-set inscription, indicates the place
current marking. Thus, the agent and the message place tokens have a $ts_{1}$ and a $ts_{2}$ time stamps
respectively. The $ts_{1}$ time stamp denotes the time by which $agent_{2}$ has received the $msg_{1}$
communicative act sent by $agent_{1}$. The $ts_{2}$ time stamp indicates the time by which $agent_{2}$ is ready
to send $msg_{2}$ response message to $agent_{1}$. Thus, the $agent_{2}$ response time $t_{2}$
(Figure~\ref{fig-duration-msg}-a) is $ts_{2}-ts_{1}$.

%-------------------------------------------------------------------------
\section{Algorithm and a Concluding Example}
\label{algorithm}

Our final contribution in this paper is a skeleton procedure for transforming an AUML conversation protocol
diagram of two interacting agents to its CP-net representation. The procedure is semi-automated--it relies on
the human to fill in some details--but also has automated aspects. We apply this procedure on a complex
multi-agent conversation protocol that involves many of the interaction building blocks already discussed.

The procedure is shown in Algorithm~\ref{alg-conversion}. The algorithm input is an AUML protocol diagram and
the algorithm creates, as an output, a corresponding CP-net representation. The CP-net is constructed in
iterations using a queue. The algorithm essentially creates the conversation net by exploring the interaction
protocol breadth-first while avoiding cycles.

\begin{algorithm}[h]
\caption{Create Conversation Net(\textbf{input}:$AUML$,\textbf{output}:$CPN$)} \label{alg-conversion}
\begin{algorithmic}[1]
\STATE $S \leftarrow \mathrm{\textbf{new}\ queue}$ \STATE $CPN \leftarrow \ \textbf{new}\
\mathrm{CP}-\mathrm{net}$ \STATE \STATE $A_{1}B_{1} \leftarrow \mathrm{\textbf{new}\ agent\ place\ with\ color\
information}$ \STATE $S.enqueue(A_{1}B_{1})$ \STATE \WHILE {$S\ \mathrm{\textbf{not}\ } empty$} \STATE $curr
\leftarrow S.dequeue()$ \STATE \STATE $MP \leftarrow CreateMessagePlaces(AUML,curr)$ \STATE $RP \leftarrow
CreateResultingAgentPlaces(AUML,curr,MP)$ \STATE $(TR,AR) \leftarrow CreateTransitionsAndArcs(AUML,curr,MP,RP)$
\STATE $FixColor(AUML,CPN,MP,RP,TR,AR)$ \STATE \FOR {$\mathrm{\textbf{each}}\ place\ p\ \mathrm{\textbf{in}}\
RP$} \IF {$p\ \mathrm{was\ not\ created\ in\ current\ iteration}$} \STATE \textbf{continue} \ENDIF \IF {$p
\mathrm{\ \textbf{is}\ \textbf{not}\ terminating\ place}$} \STATE $S.enqueue(p)$ \ENDIF \ENDFOR \STATE \STATE
$CPN.places = CPN.places \bigcup MP \bigcup RP$ \STATE $CP.transitions = CPN.transitions \bigcup TR$ \STATE
$CPN.arcs = CPN.arcs \bigcup AR$ \ENDWHILE \STATE \STATE $\mathrm{\textbf{return}}\ CPN$
\end{algorithmic}
\end{algorithm}

Lines 1-2 create and initiate the algorithm queue, and the output CP-net, respectively. The queue, denoted by
$S$, holds the initiating agent places of the current iteration. These places correspond to interaction states
that initiate further conversation between the interacting agents. In lines 4-5, an initial agent place
$A_{1}B_{1}$ is created and inserted into the queue. The $A_{1}B_{1}$ place represents a joint initial
interaction state for the two agents. Lines 7-23 contain the main loop.

We enter the main loop in line 8 and set the $curr$ variable to the first initiating agent place in $S$ queue.
Lines 10-13 create the CP-net components corresponding to the current iteration as follows. First, in line 10,
message places, associated with $curr$ agent place, are created using the $CreateMessagePlaces$ procedure (which
we do not detail here). This procedure extracts the communicative acts that are associated with a given
interaction state, from the AUML diagram. These places correspond to communicative acts, which take agents from
the joint interaction state $curr$ to its successor(s). Then in line 11, the $CreateResultingAgentPlaces$
procedure creates agent places that correspond to interaction state changes as a result of the communicative
acts associated with $curr$ agent place (again based on the AUML diagram). Then, in $CreateTransitionsAndArcs$
procedure (line 12), these places are connected using the principles described in
Sections~\ref{simple-and-complex-building-blocks}--\ref{temporal-aspects}. Thus, the CP-net structure (net
places, transitions and arcs) is created. Finally, in line 13, the $FixColor$ procedure adds token color
elements to the CP-net structure, to support deadlines, cardinality, and other communicative act attributes.

Lines 15-19 determine which resulting agent places are inserted into the $S$ queue for further iteration. Only
non-terminating agent places, i.e. places that do not correspond to interaction states that terminate the
interaction, are inserted into the queue in lines 18-19. However, there is one exception (lines 16-17): a
resulting agent place, which has already been handled by the algorithm, is not inserted back into the $S$ queue
since inserting it can cause an infinite loop. Thereafter, completing the current iteration, the output CP-net,
denoted by $CPN$ variable, is updated according to the current iteration CP-net components in lines 21-23. This
main loop iterates as long as the $S$ queue is not empty. The resulting CP-net is returned--line 25.

\begin{figure}[ht]
  \begin{center}

\includegraphics[width=0.52\columnwidth,keepaspectratio]{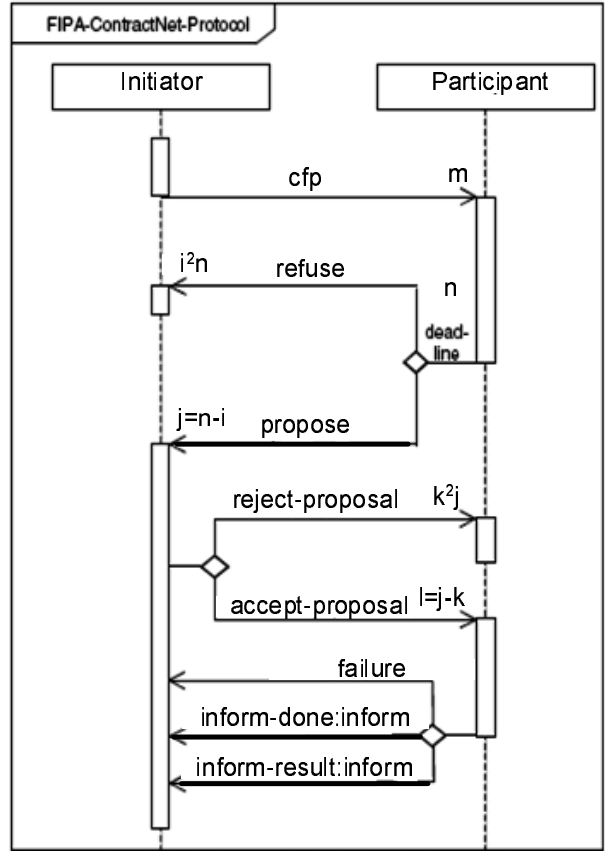}\\
  \caption {FIPA Contract Net Interaction Protocol using
AUML.}\label{fig-contract-net-auml} \vspace{-15pt}
  \end{center}
\end{figure}

To demonstrate this algorithm, we will now use it on the \emph{FIPA Contract Net Interaction Protocol}
\cite{fipac} (Figure~\ref{fig-contract-net-auml}). This protocol allows interacting agents to negotiate. The
$Initiator$ agent issues $m$ calls for proposals using a $cfp$ communicative act. Each of the $m$ $Participants$
may refuse or counter-propose by a given $deadline$ sending either a $refuse$ or a $propose$ message
respectively. A $refuse$ message terminates the interaction. In contrast, a $propose$ message continues the
corresponding interaction.

Once the $deadline$ expires, the $Initiator$ does not accept any further $Participant$ response messages. It
evaluates the received $Participant$ proposals and selects one, several, or no agents to perform the requested
task. Accepted proposal result in the sending of \emph{accept-proposal} messages, while the remaining proposals
are rejected using \emph{reject-proposal} message. \emph{Reject-proposal} terminates the interaction with the
corresponding $Participant$. On the other hand, the \emph{accept-proposal} message commits a $Participant$ to
perform the requested task. On successful completion, $Participant$ informs $Initiator$ sending either an
\emph{inform-done} or an \emph{inform-result} communicative act. However, in case a $Participant$ has failed to
accomplish the task, it communicates a $failure$ message.

\begin{figure}[ht]
  \begin{center}

\includegraphics[width=0.90\columnwidth,keepaspectratio]{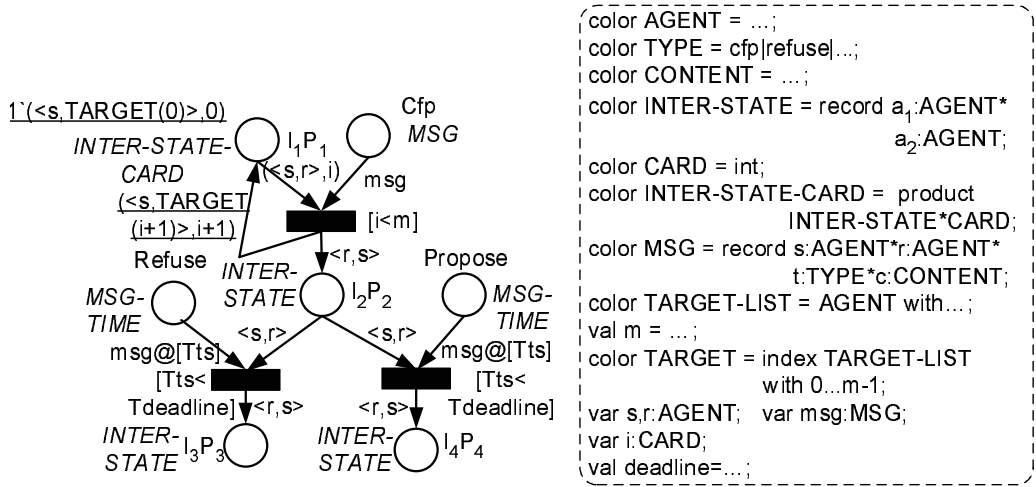}\\
  \caption {FIPA Contract Net Interaction Protocol using CP-net after
the $2^{nd}$ iteration.}\label{fig-contract-net-cp-2nd} \vspace{-15pt}
  \end{center}
\end{figure}

\begin{figure}[ht]
  \begin{center}

\includegraphics[width=0.90\columnwidth,keepaspectratio]{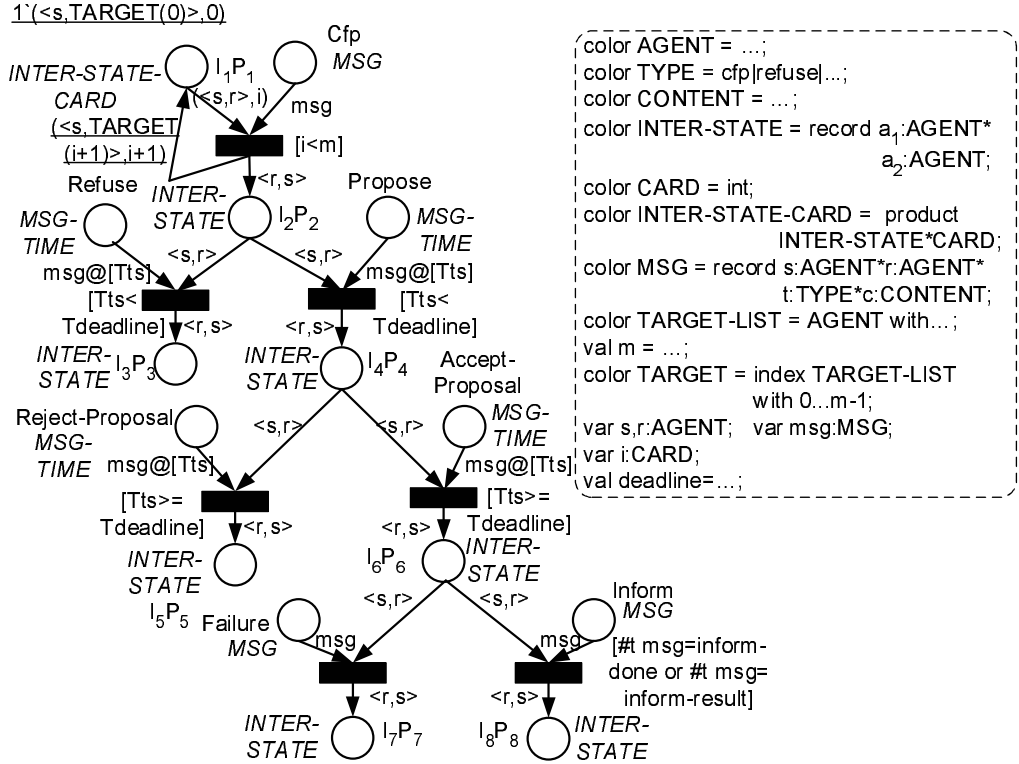}\\
  \caption {FIPA Contract Net Interaction Protocol using CP-net after
the $4^{th}$ (and final) iteration.}\label{fig-contract-net-cp-4th} \vspace{-25pt}
  \end{center}
\end{figure}

We now use the algorithm introduced above to create a CP-net, which represents the \emph{FIPA Contract Net
Interaction Protocol}. The corresponding CP-net model is constructed in four iterations of the algorithm.
Figure~\ref{fig-contract-net-cp-2nd} shows the CP-net representation after the second iteration of the
algorithm, while Figure~\ref{fig-contract-net-cp-4th} shows the CP-net representation after the fourth and final
iteration.

The \emph{Contract Net Interaction Protocol} starts from $I_{1}P_{1}$ place, which represents a joint
interaction state where $Initiator$ is ready to send a $cfp$ communicative act ($I_{1}$) and $Participant$ is
waiting for the corresponding $cfp$ message ($P_{1}$). The $I_{1}P_{1}$ place is created and inserted into the
queue before the iterations through the main loop begin.

\textbf{First iteration}. The $curr$ variable is set to the $I_{1}P_{1}$ place. The algorithm creates net
places, which are associated with the $I_{1}P_{1}$ place, i.e. a $Cfp$ message place, and an $I_{2}P_{2}$
resulting agent place. The $I_{2}P_{2}$ place denotes an interaction state in which $Initiator$ has already sent
a $cfp$ communicative act to $Participant$ and is now waiting for its response ($I_{2}$) and $Participant$ has
received the $cfp$ message and is now deciding on an appropriate response ($P_{2}$). These are created using the
$CreateMessagePlaces$ and the $CreateResultingAgentPlaces$ procedures, respectively.

Then, the $CreateTransitionsAndArcs$ procedure in line 12, connects the three places using a simple asynchronous
message building block as shown in Figure~\ref{fig-asynch-msg}-b
(Section~\ref{simple-and-complex-building-blocks}). In line 13, as the color sets of the places are determined,
the algorithm also handles the cardinality of the $cfp$ communicative act, by putting an appropriate sequence
expression on the transition, using the principles presented in Figure~\ref{fig-broadcast-msg}-b (Section
~\ref{interaction-attributes}). Accordingly, the color set, associated with $I_{1}P_{1}$ place, is changed to
the \emph{INTER-STATE-CARD} color set. Since the $I_{2}P_{2}$ place is not a terminating place, it is inserted
into the $S$ queue.

\textbf{Second iteration}.  $curr$ is set to the $I_{2}P_{2}$ place. The $Participant$ agent can send, as a
response, either a $refuse$ or a $propose$ communicative act. $Refuse$ and $Propose$ message places are created
by $CreateMessagePlaces$ (line 10), and resulting places $I_{3}P_{3}$ and $I_{4}P_{4}$, corresponding to the
results of the $refuse$ and $propose$ communicative acts, respectively, are created by
$CreateResultingAgentPlaces$ (line 11). The $I_{3}P_{3}$ place represents a joint interaction state where
$Participant$ has sent the $refuse$ message and terminated ($P_{3}$), while $Initiator$ has received it, and
terminated ($I_{3}$). The $I_{4}P_{4}$ place represents the joint state in which $Participant$ has sent the
$propose$ message ($P_{4}$), while $Initiator$ has received the message and is considering its response
($I_{4}$).

In line 12, the $I_{2}P_{2}$, $Refuse$, $I_{3}P_{3}$, $Propose$ and $I_{4}P_{4}$ places are connected using the
XOR-decision building block presented in Figure~\ref{fig-xor-msg}-b
(Section~\ref{simple-and-complex-building-blocks}). Then, the $FixColor$ procedure (line 13), adds the
appropriate token color attributes, to allow a deadline sequence expression (on both the $refuse$ and the
$propose$ messages) to be implemented as shown in Figure~\ref{fig-deadline-msg}-b
(Section~\ref{temporal-aspects}). The $I_{3}P_{3}$ place denotes a terminating state, whereas the $I_{4}P_{4}$
place continues the interaction. Thus, in lines 18-19, only the $I_{4}P_{4}$ place is inserted into the queue,
for the next iteration of the algorithm. The state of the net at the end of the second iteration of the
algorithm is presented in Figure~\ref{fig-contract-net-cp-2nd}.

\textbf{Third iteration}. $curr$ is set to $I_{4}P_{4}$. Here, the $Initiator$ response to a $Participant$
proposal can either be an \emph{accept-proposal} or a \emph{reject-proposal}. $CreateMessagePlaces$ procedure in
line 10 thus creates the corresponding \emph{Accept-Proposal} and \emph{Reject-Proposal} message places. The
\emph{accept-proposal} and \emph{reject-proposal} messages cause the interacting agents to transition to
$I_{5}P_{5}$ and $I_{6}P_{6}$ places, respectively. These agent places are created using the
$CreateResultingAgentPlaces$ procedure (line 11). The $I_{5}P_{5}$ place denotes an interaction state in which
$Initiator$ has sent a \emph{reject-proposal} message and terminated the interaction ($I_{5}$), while the
$Participant$ has received the message and terminated as well ($P_{5}$). In contrast, the $I_{6}P_{6}$ place
represents an interaction state where $Initiator$ has sent an \emph{accept-proposal} message and is waiting for
a response ($I_{6}$), while $Participant$ has received the \emph{accept-proposal} communicative act and is now
performing the requested task before sending a response ($P_{6}$). The $Initiator$ agent sends exclusively
either an \emph{accept-proposal} or a \emph{reject-proposal} message. Thus, the $I_{4}P_{4}$,
\emph{Reject-Proposal}, $I_{5}P_{5}$, \emph{Accept-Proposal} and $I_{6}P_{6}$ places are connected using a
XOR-decision block (in the $CreateTransitionsAndArcs$ procedure, line 12).

The $FixColor$ procedure in line 13 operates now as follows: According to the interaction protocol semantics,
the $Initiator$ agent evaluates all the received $Participant$ proposals once the $deadline$ passes. Only
thereafter, the appropriate \emph{reject-proposal} and \emph{accept-proposal} communicative acts are sent. Thus,
$FixColor$ assigns a \emph{MSG-TIME} color set to the \emph{Reject-Proposal} and the \emph{Accept-Proposal}
message places, and creates a $[Tts>=Tdeadline]$ transition guard on the associated transitions. This transition
guard guarantees that $Initiator$ cannot send any response until the $deadline$ expires, and all valid
$Participant$ responses have been received. The resulting $I_{5}P_{5}$ agent place denotes a terminating
interaction state, whereas the $I_{6}P_{6}$ agent place continues the interaction. Thus, only $I_{6}P_{6}$ agent
place is inserted into the $S$ queue.

\textbf{Fourth iteration}.  $curr$ is set to $I_{6}P_{6}$. This place is associated with three communicative
acts: \emph{inform-done}, \emph{inform-result} and $failure$. The \emph{inform-done} and the
\emph{inform-result} messages are instances of the $inform$ communicative act class. Thus,
\emph{CreateMessagePlaces} (line 10) creates only two message places, $Inform$ and $Failure$. In line 11,
$CreateResultingAgentPlaces$ creates the $I_{7}P_{7}$ and $I_{8}P_{8}$ agent places. The $failure$ communicative
act causes interacting agents to transition to $I_{7}P_{7}$ agent place, while both $inform$ messages cause the
agents to transition to $I_{8}P_{8}$ agent place. The $I_{7}P_{7}$ place represents a joint interaction state
where $Participant$ has sent the $failure$ message and terminated ($P_{7}$), while $Initiator$ has received a
$failure$ communicative act and terminated ($I_{7}$). On the other hand, the $I_{8}P_{8}$ place denotes an
interaction state in which $Participant$ has sent the $inform$ message (either \emph{inform-done} or
\emph{inform-result}) and terminated ($P_{8}$), while $Initiator$ has received an $inform$ communicative act and
terminated ($I_{8}$). The $inform$ and $failure$ communicative acts are sent exclusively. Thus
$CreateTransitionsAndArcs$ (line 12) connects the $I_{6}P_{6}$, $Failure$, $I_{7}P_{7}$, $Inform$ and
$I_{8}P_{8}$ places using a XOR-decision building block. Then, $FixColor$ assigns a $[\#t\
msg=\mathrm{\emph{inform-done}}\ \mathrm{or}\ \#t\ msg=\mathrm{\emph{inform-result}}]$ transition guard on the
transition associated with $Inform$ message place. Since both the $I_{7}P_{7}$ and the $I_{8}P_{8}$ agent places
represent terminating interaction states, they are not inserted into the queue, which remains empty at the end
of the current iteration. This signifies the end of the conversion. The complete conversation CP-net resulting
after this iteration of the algorithm is shown in Figure~\ref{fig-contract-net-cp-4th}.

The procedure we outline can guide the conversion of many \emph{2-agent} conversation protocols in AUML to their
CP-net equivalents. However, it is not sufficiently developed to address the general \emph{n-agent} case.
Appendix~\ref{complex-example} presents a complex example of a \emph{3-agent} conversation protocol, which was
successfully converted manually, without the guidance of the algorithm. This example incorporates many advanced
features of our CP-net representation technique and would have been beyond the scope of many previous
investigations.

%-------------------------------------------------------------------------
\section{Summary and Conclusions}
\label{summary}

Over recent years, open distributed MAS applications have gained broad acceptance both in the multi-agent
academic community and in real-world industry. As a result, increasing attention has been directed to
multi-agent conversation representation techniques. In particular, Petri nets have recently been shown to
provide a viable representation approach \cite{cost99b,cost00,nowostawski01,mazouzi02}.

However, radically different approaches have been proposed to using Petri nets for modelling multi-agent
conversations. Yet, the relative strengths and weaknesses of the proposed techniques have not been examined. Our
work introduces a novel classification of previous investigations and then compares these investigations
addressing their scalability and appropriateness for overhearing tasks.

Based on the insights gained from the analysis, we have developed a novel representation, that uses CP-nets in
which places explicitly represent joint interaction states and messages. This representation technique offers
significant improvements (compared to previous approaches) in terms of scalability, and is particularly suitable
for monitoring via overhearing. We systematically show how this representation covers essentially all the
features required to model complex multi-agent conversations, as defined by the FIPA conversation standards
\cite{fipaa}. These include simple \& complex interaction building blocks
(Section~\ref{simple-and-complex-building-blocks} \& Appendix~\ref{additional-examples}), communicative act
attributes and multiple concurrent conversations using the same CP-net (Section~\ref{interaction-attributes}),
nested \& interleaved interactions using hierarchical CP-nets
(Section~\ref{nested-and-interleaved-interactions}) and temporal interaction attributes using timed CP-nets
(Section~\ref{temporal-aspects}). The developed techniques have been demonstrated, throughout the paper, on
complex interaction protocols defined in the FIPA conversation standards (see in particular the example
presented in Appendix~\ref{complex-example}). Previous approaches could handle some of these examples (though
with reduced scalability), but only a few were shown to cover all the required features.

Finally, the paper presented a skeleton procedure for semi-automatically converting an AUML protocol diagrams
(the chosen FIPA representation standard) to an equivalent CP-net representation. We have demonstrated its use
on a challenging FIPA conversation protocol, which was difficult to represent using previous approaches.

We believe that this work can assist and motivate continuing research on multi-agent conversations including
such issues as performance analysis, validation and verification \cite{desel97}, agent conversation
visualization, automated monitoring \cite{kaminka02,busetta01,busetta02}, deadlock detection \cite{khomenko00},
debugging \cite{poutakidis02} and dynamic interpretation of interaction protocols \cite{cranefield02,desilva03}.
Naturally, some issues remain open for future work. For example, the presented procedure addresses only AUML
protocol diagrams representing two agent roles. We plan to investigate an \emph{n-agent} version in the future.

\acks{The authors would like to thank the anonymous JAIR reviewers for many useful and informative comments.
Minor subsets of this work were also published as LNAI book chapter \cite{gutnik04b}. K. Ushi deserves many
thanks. }

%-------------------------------------------------------------------------
\appendix
\section{A Brief Introduction to Petri Nets}
\label{petri-nets-introduction}

\begin{figure}[ht]
\begin{center}
\subfigure[Before firing]{
\includegraphics[width=0.15\columnwidth,keepaspectratio]{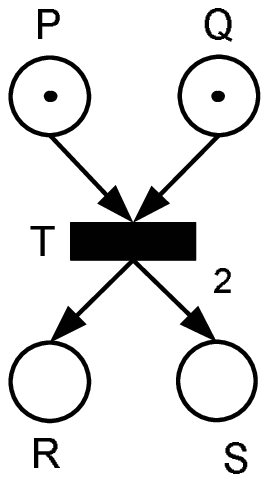}}
\hspace{0.02\columnwidth} \subfigure[After firing]{
\includegraphics[width=0.15\columnwidth,keepaspectratio]{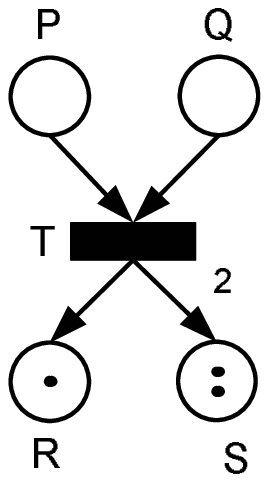}}
\caption {A PT-net example.}
  \label{fig-pt-nets}
\vspace{-15pt}
  \end{center}
\end{figure}

Petri nets \cite{petriNets} are a widespread, established methodology for representing and reasoning about
distributed systems, combining a graphical representation with a comprehensive mathematical theory. One version
of Petri nets is called Place/Transition nets (PT-nets) \cite{reisig85}. A PT-net is a bipartite directed graph
where each node is either a place or a transition (Figure~\ref{fig-pt-nets}). The net places and transitions are
indicated through circles and rectangles respectively. The PT-net arcs support only
$\mathrm{place}\rightarrow{\mathrm{transition}}$ and $\mathrm{transition}\rightarrow{\mathrm{place}}$
connections, but never connections between two places or between two transitions. The arc direction determines
the input/output characteristics of the place and the transition connected. Thus, given an arc,
$P\rightarrow{T}$, connecting place $P$ and transition $T$, we will say that place $P$ is an input place of
transition $T$ and vice versa transition $T$ is an output transition of place $P$. The $P\rightarrow{T}$ arc is
considered to be an output arc of place $P$ and an input arc of transition $T$.

A PT-net place may be \emph{marked} by small black dots called \emph{tokens}. The arc expression is an integer,
which determines the number of tokens associated with the corresponding arc. By convention, an arc expression
equal to 1 is omitted. A specific transition is \emph{enabled} if and only if its input places \emph{marking}
satisfies the appropriate arc expressions. For example, consider arc $P \rightarrow T$ to be the only arc to
connect place $P$ and transition $T$. Thus, given that this arc has an arc expression 2, we will say that
transition $T$ is enabled if and only if place $P$ is marked with two tokens. In case the transition is enabled,
it may \emph{fire/occur}. The transition occurrence removes tokens from the transition input places and puts
tokens to the transition output places as specified by the arc expressions of the corresponding input/output
arcs. Thus, in Figures~\ref{fig-pt-nets}-a and \ref{fig-pt-nets}-b, we demonstrate PT-net marking before and
after transition firing respectively.

Although computationally equivalent, a different version of Petri nets, called \emph{Colored Petri nets}
(CP-nets) \cite{jensen97a,jensen97b,jensen97c}, offers greater flexibility in compactly representing complex
systems. Similarly to the PT-net model, CP-nets consist of net places, net transitions and arcs connecting them.
However, in CP-nets, tokens are not just single bits, but can be complex, structured, information carriers. The
type of additional information carried by the token, is called \emph{token color}, and it may be simple (e.g.,
an integer or a string), or complex (e.g. a record or a tuple). Each place is declared by a \emph{place color}
set to only match tokens of particular colors. A CP-net place marking is a token \emph{multi-set} (i.e., a set
in which a member may appear more than once) corresponding to the appropriate place color set. CP-net arcs pass
token multi-sets between the places and transitions. CP-net arc expressions can evaluate token multi-sets and
may involve complex calculation procedures over token \emph{variables} declared to be associated with the
corresponding arcs.

The CP-net model introduces additional extensions to PT-nets. \emph{Transition guards} are boolean expressions,
which constrain transition firings. A transition guard associated with a transition tests tokens that pass
through a transition, and will only enable the transition firings if the guard is successfully matched (i.e.,
the test evaluates to true). The CP-net transition guards, together with places color sets and arc expressions,
appear as a part of net \emph{inscriptions} in the CP-net.

In order to visualize and manage the complexity of large CP-nets, hierarchical CP-nets \cite{huber91,jensen97a}
allow hierarchical representations of CP-nets, in which sub-CP nets can be re-used in higher-level CP nets, or
abstracted away from them. Hierarchical CP-nets are built from pages, which are themselves CP-nets.
\emph{Superpages} present a higher level of hierarchy, and are CP-nets that refer to \emph{subpages}, in
addition to transitions and places. A subpage may also function as a superpage to other subpages. This way,
multiple hierarchy levels can be used in a hierarchical CP-net structure.

The relationship between a superpage and a subpage is defined by a \emph{substitution transition}, which
substitutes a corresponding \emph{subpage instance} on the CP-net superpage structure as a transition in the
superpage. The substitution transition \emph{hierarchy inscription} supplies the exact mapping of the superpage
places connected to the substitution transition (called \emph{socket nodes}), to the subpage places (called
\emph{port nodes}). The \emph{port types} determine the characteristics of the socket node to port node
mappings. A complete CP-net hierarchical structure is presented using a \emph{page hierarchy graph}, a directed
graph where vertices correspond to pages, and directed edges correspond to direct superpage-subpage
relationships.

\emph{Timed CP-nets} \cite{jensen97b} extend CP-nets to support the representation of temporal aspects using a
\emph{global clock}. Timed CP-net tokens have an additional color attribute called \emph{time stamp}, which
refers to the earliest time at which the token may be used. Time stamps can be used by arc expression and
transition guards, to enable a timed-transition if and only if it satisfies two conditions: (i) the transition
is \emph{color enabled}, i.e. it satisfies the constraints defined by arc expression and transition guards; and
(ii) the tokens are \emph{ready}, i.e. the time of the global clock is equal to or greater than the tokens' time
stamps. Only then can the transition fire.

%-------------------------------------------------------------------------
\section{Additional Examples of Conversation Representation Building
Blocks} \label{additional-examples}

\begin{figure}[ht]
\begin{center}
\subfigure[AUML representation]{
\includegraphics[width=0.36\columnwidth,keepaspectratio]{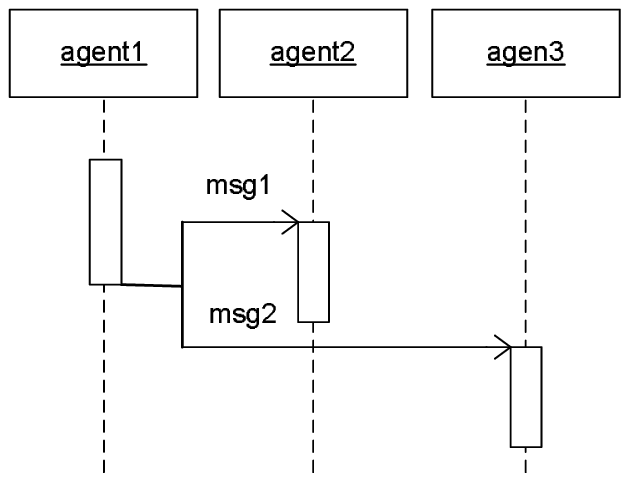}}
\hspace{0.005\columnwidth} \subfigure[CP-net representation]{
\includegraphics[width=0.60\columnwidth,keepaspectratio]{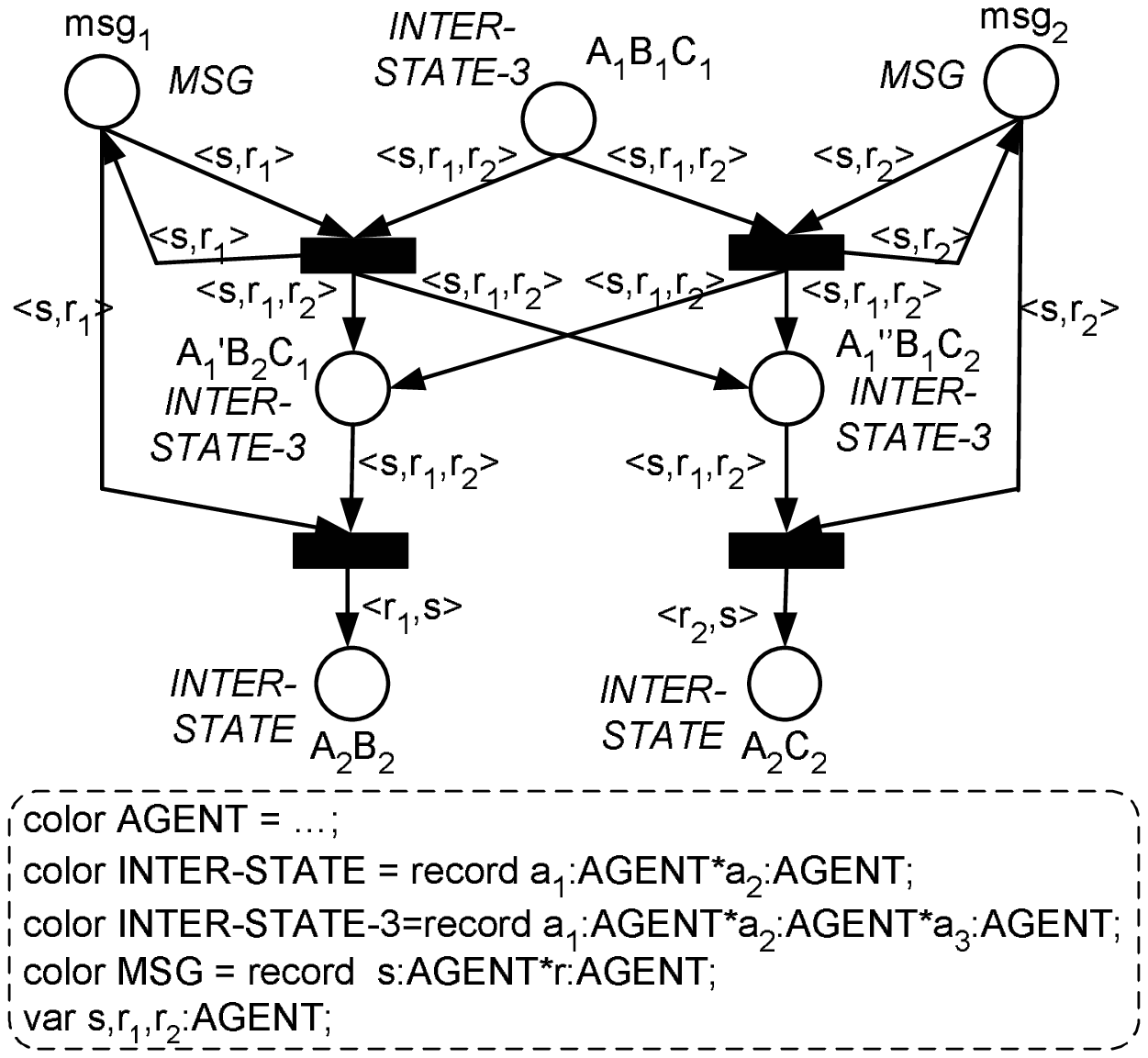}}
\caption {AND-parallel messages interaction.}
  \label{fig-and-parallel-msg}
\vspace{-25pt}
  \end{center}
\end{figure}

This appendix presents some additional interaction building blocks to those already described in
Section~\ref{simple-and-complex-building-blocks}. The first is the AND-parallel messages interaction (AUML
representation shown in Figure~\ref{fig-and-parallel-msg}-a). Here, the sender $agent_{1}$ sends both the
$msg_{1}$ message to $agent_{2}$ and the $msg_{2}$ message to $agent_{3}$. However, the order of the two
communicative acts is unconstrained.

The representation of AND-parallel in our CP-net representation is shown in Figure~\ref{fig-and-parallel-msg}-b.
The $A_{1}B_{1}C_{1}$, $A_{2}B_{2}$, $A_{2}C_{2}$, $msg_{1}$ and $msg_{2}$ places are defined similarly to
Figures~\ref{fig-xor-msg}-b and \ref{fig-or-parallel-msg}-b in Section~\ref{simple-and-complex-building-blocks}.
However, we also define two additional intermediate agent places, $A_{1}'B_{2}C_{1}$ and $A_{1}''B_{1}C_{2}$.
The $A_{1}'B_{2}C_{1}$ place represents a joint interaction state where $agent_{1}$ has sent the $msg_{1}$
message to $agent_{2}$ and is ready to send the $msg_{2}$ communicative act to $agent_{3}$ ($A_{1}$'),
$agent_{2}$ has received the $msg_{1}$ message ($B_{2}$) and $agent_{3}$ is waiting to receive the $msg_{2}$
communicative act ($C_{1}$). The $A_{1}''B_{1}C_{2}$ place represents a joint interaction state in which
$agent_{1}$ is ready to send the $msg_{1}$ message to $agent_{2}$ and has already sent the $msg_{2}$
communicative act to $agent_{3}$ ($A_{1}''$), $agent_{2}$ is waiting to receive the $msg_{1}$ message ($B_{1}$)
and $agent_{3}$ has received the $msg_{2}$ communicative act ($C_{2}$). These places enable $agent_{1}$ to send
both communicative acts concurrently. Four transitions connect the appropriate places respectively. The behavior
of the transitions connecting $A_{1}'B_{2}C_{1} \rightarrow A_{2}B_{2}$ and $A_{1}''B_{1}C_{2} \rightarrow
A_{2}C_{2}$ is similar to described above. The transitions $A_{1}B_{1}C_{1} \rightarrow A_{1}'B_{2}C_{1}$ and
$A_{1}B_{1}C_{1} \rightarrow A_{1}''B_{1}C_{2}$ are triggered by receiving messages $msg_{1}$ and $msg_{2}$,
respectively. However, these transitions should not consume the message token since it is used further for
triggering transitions $A_{1}'B_{2}C_{1} \rightarrow A_{2}B_{2}$ and $A_{1}''B_{1}C_{2} \rightarrow A_{2}C_{2}$.
This is achieved by adding an appropriate message place as an output place of the corresponding transition.

\begin{figure}[ht]
\begin{center}
\subfigure[AUML representation]{
\includegraphics[width=0.35\columnwidth,keepaspectratio]{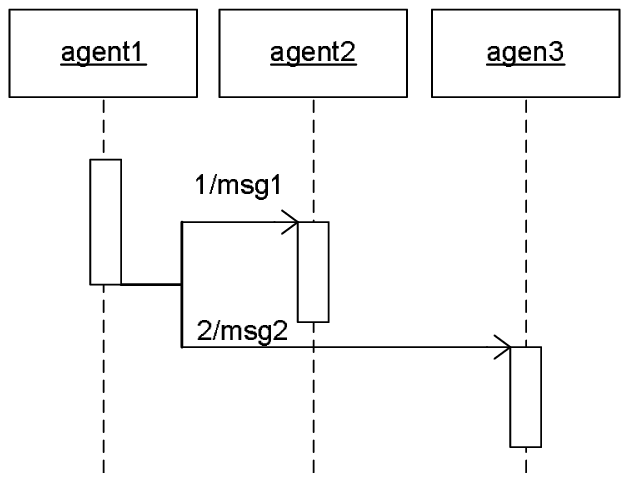}}
\hspace{0.005\columnwidth} \subfigure[CP-net representation]{
\includegraphics[width=0.61\columnwidth,keepaspectratio]{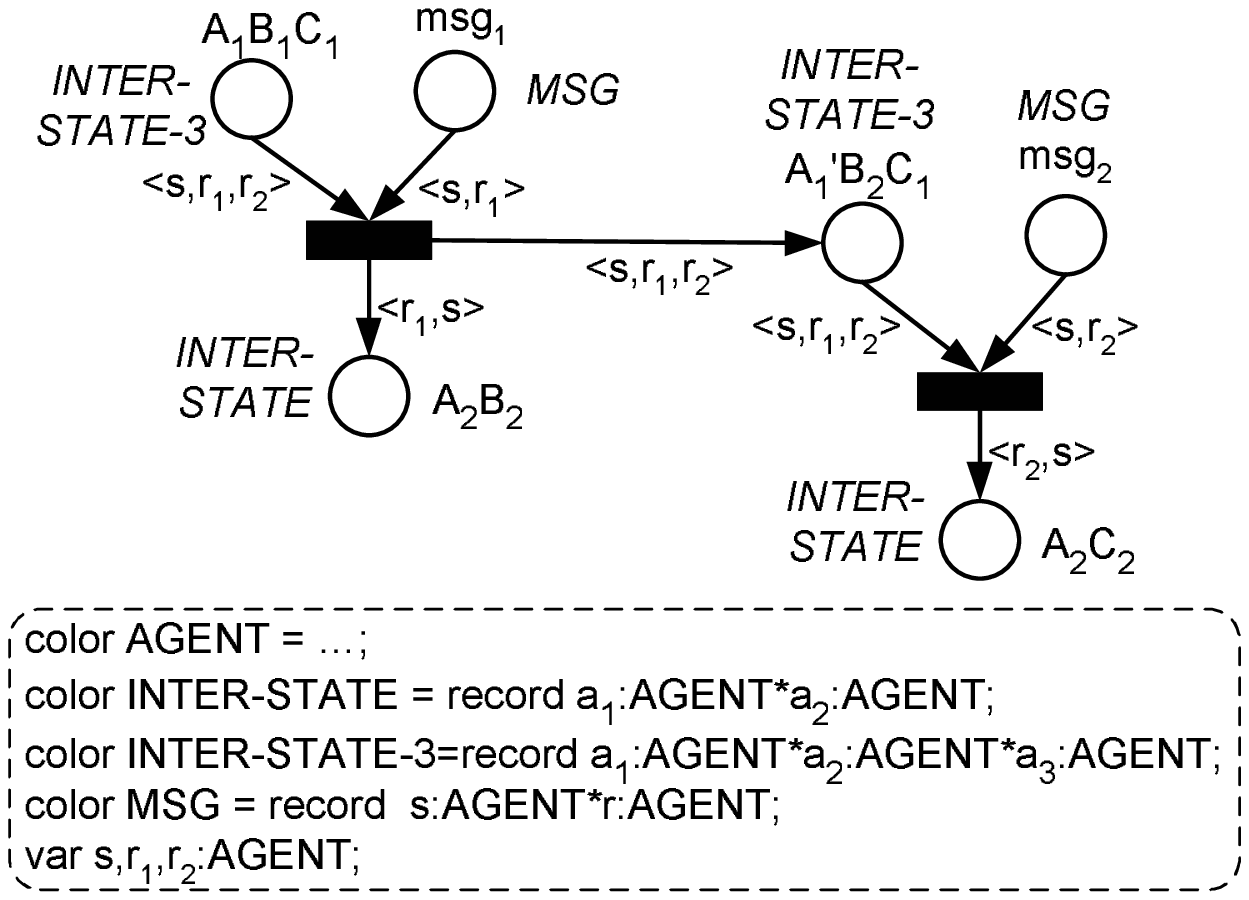}}
\caption {Sequence messages interaction.}
  \label{fig-sequence-msg}
\vspace{-20pt}
  \end{center}
\end{figure}

The second AUML interaction building block, shown in Figure~\ref{fig-sequence-msg}-a, is the message sequence
interaction, which is similar to AND-parallel. However, the message sequence interaction defines explicitly the
order between the transmitted messages. Using the $1/msg_{1}$ and $2/msg_{2}$ notation,
Figure~\ref{fig-sequence-msg}-a specifies that the $msg_{1}$ message should be sent before sending $msg_{2}$.

Figure~\ref{fig-sequence-msg}-b shows the corresponding CP-net representation. The $A_{1}B_{1}C_{1}$,
$A_{2}B_{2}$, $A_{2}C_{2}$, $msg_{1}$ and $msg_{2}$ places are defined as before. However, the CP-net
implementation presents an additional intermediate agent place--$A_{1}'B_{2}C{1}$--which is identical to the
corresponding intermediate agent place in Figure~\ref{fig-and-parallel-msg}-b. $A_{1}'B_{2}C_{1}$ is defined as
an output place of the $A_{1}B_{1}C_{1} \rightarrow A_{2}B_{2}$ transition. It thus guarantees that the
$msg_{2}$ communicative act can be sent (represented by the $A_{1}'B_{2}C_{1} \rightarrow A_{2}C_{2}$
transition) only upon completion of the $msg_{1}$ transmission (the $A_{1}B_{1}C_{1} \rightarrow A_{2}B_{2}$
transition).

\begin{figure}[ht]
\begin{center}
\subfigure[AUML representation]{
\includegraphics[width=0.35\columnwidth,keepaspectratio]{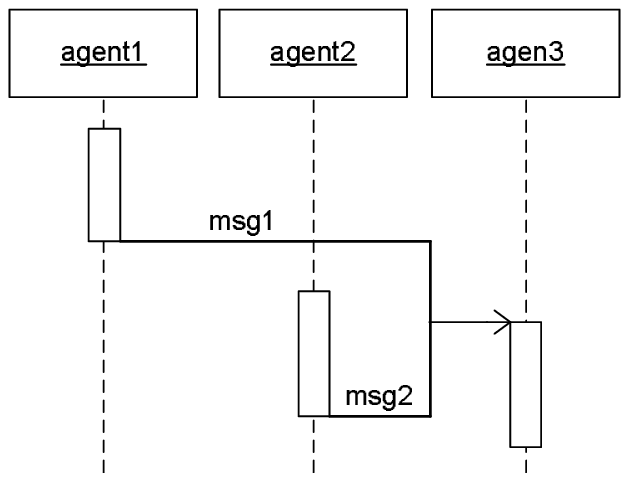}}
\hspace{0.005\columnwidth} \subfigure[CP-net representation]{
\includegraphics[width=0.61\columnwidth,keepaspectratio]{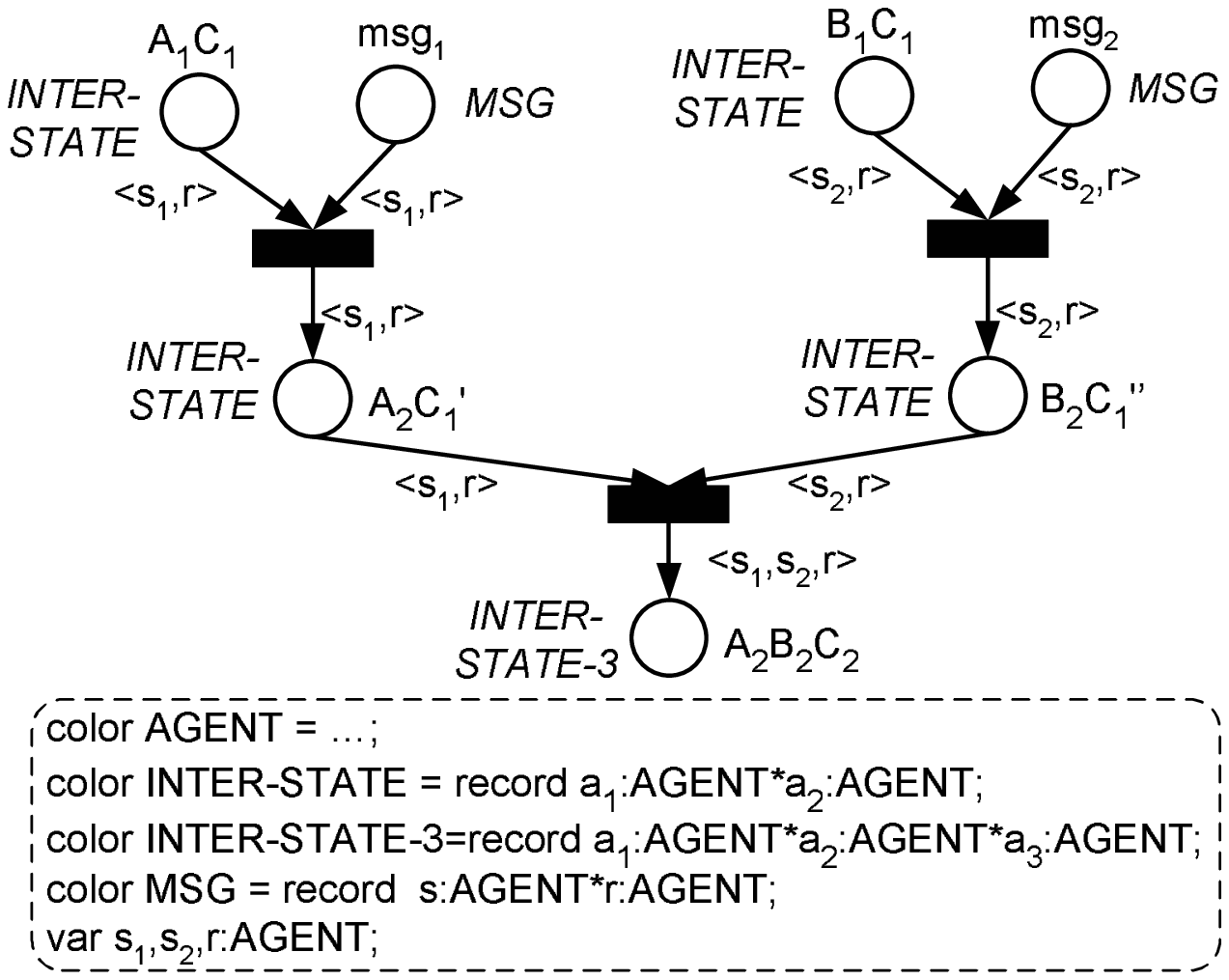}}
\caption {Synchronized messages interaction.}
  \label{fig-synchronized-msg}
\vspace{-20pt}
  \end{center}
\end{figure}

The last interaction we present is the synchronized messages interaction, shown in
Figure~\ref{fig-synchronized-msg}-a. Here, $agent_{3}$ simultaneously receives $msg_{1}$ from $agent_{1}$ and
$msg_{2}$ from $agent_{2}$. In AUML, this constraint is annotated by merging the two communicative act arrows
into a horizontal bar with a single output arrow.

Figure~\ref{fig-synchronized-msg}-b illustrates the CP-net implementation of synchronized messages interaction.
As in previous examples, we define the $A_{1}C_{1}$, $B_{1}C_{1}$, $msg_{1}$, $msg_{2}$ and $A_{2}B_{2}C_{2}$
places. We additionally define two intermediate agent places, $A_{2}C_{1}'$ and $B_{2}C_{1}''$. The
$A_{2}C_{1}'$ place represents a joint interaction state where $agent_{1}$ has sent $msg_{1}$ to $agent_{3}$
($A_{2}$), and $agent_{3}$ has received it, however $agent_{3}$ is also waiting to receive $msg_{2}$ ($C_{1}'$).
The $B_{2}C_{1}''$ place represents a joint interaction state in which $agent_{2}$ has sent $msg_{2}$ to
$agent_{3}$ ($B_{2}$), and $agent_{3}$ has received it, however $agent_{3}$ is also waiting to receive $msg_{1}$
($C_{1}''$). These places guarantee that the interaction does not transition to the $A_{2}B_{2}C_{2}$ state
until both $msg_{1}$ and $msg_{2}$ have been received by $agent_{3}$.

%-------------------------------------------------------------------------
\section{An Example of a Complex Interaction Protocol}
\label{complex-example}

\begin{figure}[ht]
  \begin{center}

\includegraphics[width=0.81\columnwidth,keepaspectratio]{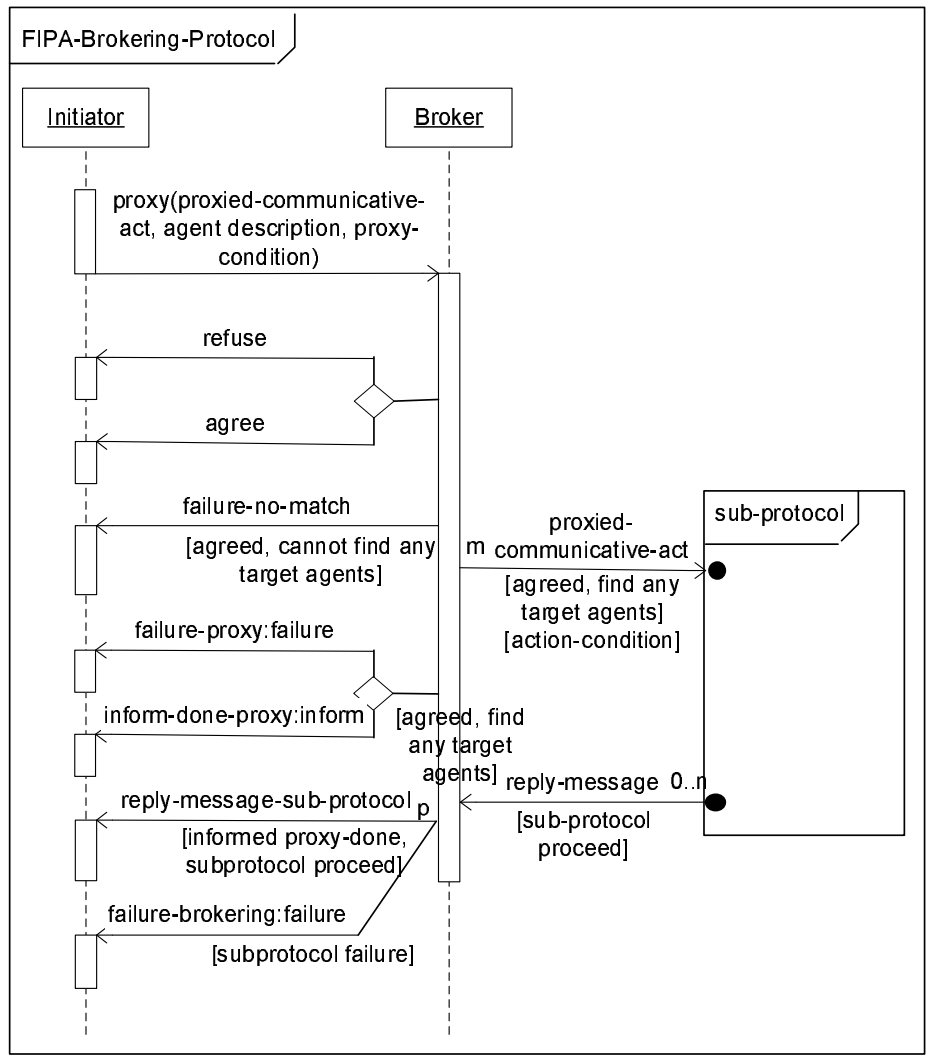}\\
  \caption {FIPA Brokering Interaction Protocol - AUML
representation.}\label{fig-brokering-auml} \vspace{-25pt}
  \end{center}
\end{figure}

We present an example of a complex \emph{3-agent} conversation protocol, which was manually converted to a
CP-net representation using the building blocks in this paper. The conversation protocol addressed here is the
\emph{FIPA Brokering Interaction Protocol} \cite{fipad}. This interaction protocol incorporates many advanced
conversation features of our representation such as nesting, communicative act sequence expression, message
guards and etc. Its AUML representation is shown in Figure~\ref{fig-brokering-auml}.

The $Initiator$ agent begins the interaction by sending a $proxy$ message to the $Broker$ agent. The $proxy$
communicative act contains the requested \emph{proxied-communicative-act} as part of its argument list. The
$Broker$ agent processes the request and responds with either an $agree$ or a $refuse$ message. Communication of
a $refuse$ message terminates the interaction. If the $Broker$ agent has agreed to function as a proxy, it then
locates the agents matching the $Initiator$ request. If no such agent can be found, the $Broker$ agent
communicates a \emph{failure-no-match} message and the interaction terminates. Otherwise, the $Broker$ agent
begins $m$ interactions with the matching agents. For each such agent, the $Broker$ informs the $Initiator$,
sending either an \emph{inform-done-proxy} or a \emph{failure-proxy} communicative act. The \emph{failure-proxy}
communicative act terminates the \emph{sub-protocol} interaction with the matching agent in question. The
\emph{inform-done-proxy} message continues the interaction. As the \emph{sub-protocol} progresses, the $Broker$
forwards the received responses to the $Initiator$ agent using the \emph{reply-message-sub-protocol}
communicative acts. However, there can be other failures that are not explicitly returned from the
\emph{sub-protocol} interaction (e.g., if the agent executing the \emph{sub-protocol} has failed). In case the
$Broker$ agent detects such a failure, it communicates a \emph{failure-brokering} message, which terminates the
\emph{sub-protocol} interaction.

\begin{figure}[ht]
  \begin{center}

\includegraphics[width=0.85\columnwidth,keepaspectratio]{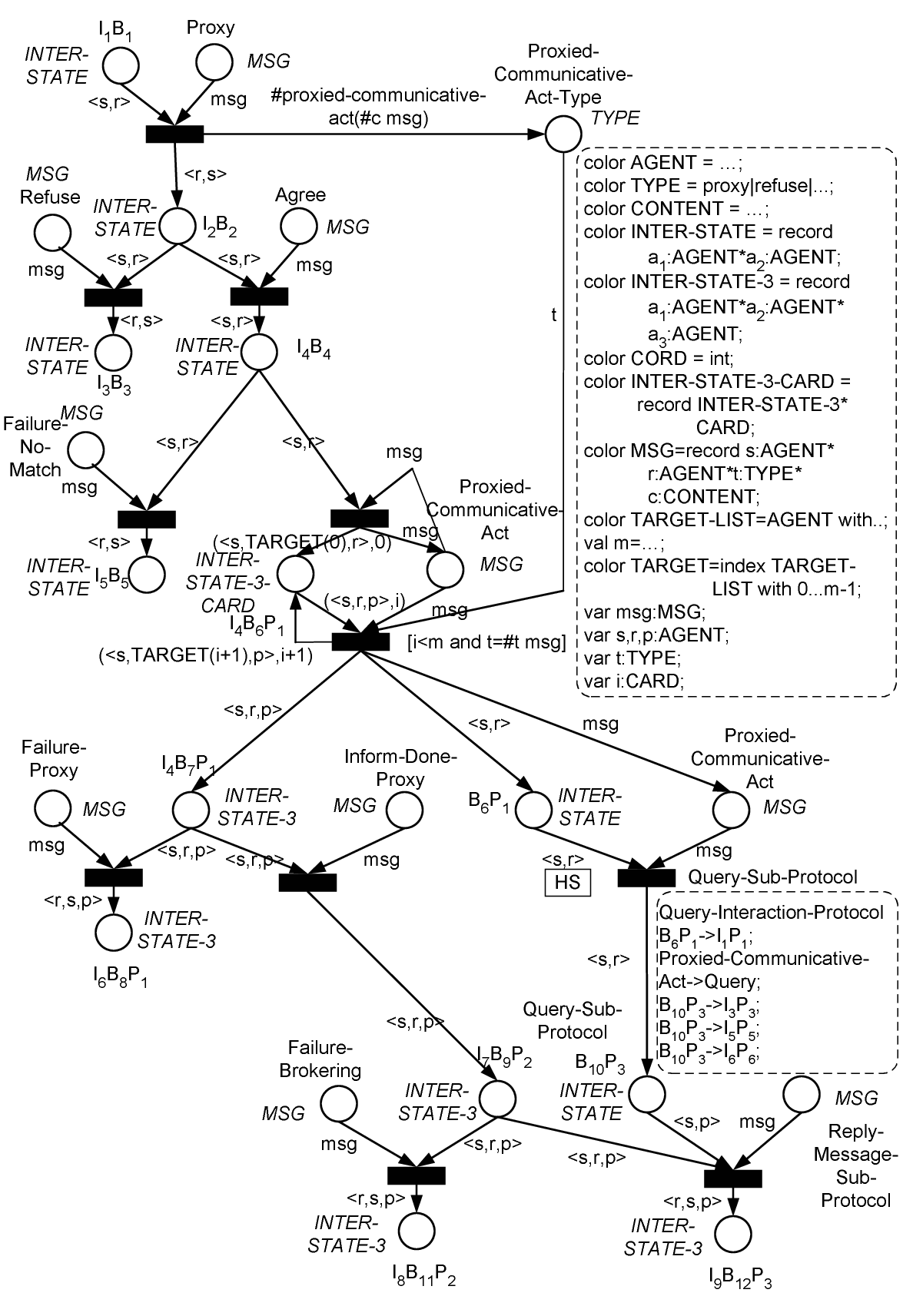}\\
  \caption {FIPA Brokering Interaction Protocol - CP-net
representation.}\label{fig-brokering-cp} \vspace{-25pt}
  \end{center}
\end{figure}

A CP-net representation of the \emph{FIPA Brokering Interaction Protocol} is shown in
Figure~\ref{fig-brokering-cp}. The \emph{Brokering Interaction Protocol} starts from $I_{1}B_{1}$ place. The
$I_{1}B_{1}$ place represents a joint interaction state where $Initiator$ is ready to send a $proxy$
communicative act ($I_{1}$) and $Broker$ is waiting to receive it ($B_{1}$). The $proxy$ communicative act
causes the interacting agents to transition to $I_{2}B_{2}$. This place denotes an interaction state in which
$Initiator$ has already sent a $proxy$ message to $Broker$ ($I_{2}$) and $Broker$ has received it ($B_{2}$). The
$Broker$ agent can send, as a response, either a $refuse$ or an $agree$ communicative act. This CP-net component
is implemented using the XOR-decision building block presented in
Section~\ref{simple-and-complex-building-blocks}. The $refuse$ message causes the agents to transition to
$I_{3}B_{3}$ place and thus terminate the interaction. This place corresponds to $Broker$ sending a $refuse$
message and terminating ($B_{3}$), while $Initiator$ receiving the message and terminating ($I_{3}$). On the
other hand, the $agree$ communicative act causes the agents to transition to $I_{4}B_{4}$ place, which
represents a joint interaction state in which the $Broker$ has sent an $agree$ message to $Initiator$ (and is
now trying to locate the receivers of the $proxied$ message), while the $Initiator$ received the $agree$
message.

The $Broker$ agent's search for suitable receivers may result in two alternatives. First, in case no matching
agents are found, the interaction terminates in the $I_{5}B_{5}$ agent place. This joint interaction place
corresponds to an interaction state where $Broker$ has sent the \emph{failure-no-match} communicative act
($B_{5}$), and $Initiator$ has received the message and terminated ($I_{5}$). The second alternative is that
suitable agents have been found. Then, $Broker$ starts sending \emph{proxied-communicative-act} messages to
these agents on the established list of designated receivers, i.e. \emph{TARGET-LIST}. The first such
\emph{proxied-communicative-act} message causes the interacting agents to transition to $I_{4}B_{6}P_{1}$ place.
The $I_{4}B_{6}P_{1}$ place denotes a joint interaction state of three agents: $Initiator$, $Broker$ and
$Participant$ (the receiver). The $Initiator$ individual state remains unchanged ($I_{4}$) since the
\emph{proxied-communicative-act} message starts an interaction between $Broker$ and $Participant$. The $Broker$
individual state ($B_{6}$) denotes that designated agents have been found and the
\emph{proxied-communicative-act} messages are ready to be sent, while $Participant$ is waiting to receive the
interaction initiating communicative act ($P_{1}$). The \emph{proxied-communicative-act} message place is also
connected as an output place of the transition. This message place is used as part of a CP-net XOR-decision
structure, which enables the $Broker$ agent to send either a \emph{failure-no-match} or a
\emph{proxied-communicative-act}, respectively. Thus, the token denoting the \emph{proxied-communicative-act}
message, must not be consumed by the transition.

Thus, multiple \emph{proxied-communicative-act} messages are sent to all $Participants$. This is implemented
similarly to the broadcast sequence expression implementation (Section~\ref{interaction-attributes}).
Furthermore, the \emph{proxied-communicative-act} type is verified against the type of the requested $proxied$
communicative act, which is obtained from the original proxy message content. We use the
\emph{Proxied-Communicative-Act-Type} message type place to implement this CP-net component similarly to
Figure~\ref{fig-query-cp}. Each \emph{proxied-communicative-act} message causes the interacting agents to
transition to both the $I_{4}B_{7}P_{1}$ and the $B_{6}P_{1}$ places.

The $B_{6}P_{1}$ place corresponds to interaction between the $Broker$ and the $Participant$ agents. It
represents a joint interaction state in which $Broker$ is ready to send a \emph{proxied-communicative-act}
message to $Participant$ ($B_{6}$), and $Participant$ is waiting for the message ($P_{1}$). In fact, the
$B_{6}P_{1}$ place initiates the nested interaction protocol that results in $B_{10}P_{3}$ place. The
$B_{10}P_{3}$ place represents a joint interaction state where $Participant$ has sent the \emph{reply-message}
communicative act and terminated ($P_{3}$), and $Broker$ has received the message ($B_{10}$). In our example, we
have chosen the \emph{FIPA Query Interaction Protocol} \cite{fipab}
(Figures~\ref{fig-query-auml}--\ref{fig-query-cp}) as the interaction \emph{sub-protocol}. The CP-net component,
implementing the nested interaction \emph{sub-protocol}, is modeled using the principles described in
Section~\ref{nested-and-interleaved-interactions}. Consequently, the interaction \emph{sub-protocol} is
concealed using the \emph{Query-Sub-Protocol} substitution transition. The $B_{6}P_{1}$,
\emph{proxied-communicative-act} and $B_{10}P_{3}$ places determine substitution transition socket nodes. These
socket nodes are assigned to the CP-net port nodes in Figure~\ref{fig-query-cp} as follows. The $B_{6}P_{1}$ and
\emph{proxied-communicative-act} places are assigned to the $I_{1}P_{1}$ and $query$ input port nodes, while the
$B_{10}P_{3}$ place is assigned to the $I_{3}P_{3}$, $I_{5}P_{5}$ and $I_{6}P_{6}$ output port nodes.

We now turn to the $I_{4}B_{7}P_{1}$ place. In contrast to the $B_{6}P_{1}$ place, this place corresponds to the
main interaction protocol. The $I_{4}B_{7}P_{1}$ place represents a joint interaction state in which $Initiator$
is waiting for $Broker$ to respond ($I_{4}$), $Broker$ is ready to send an appropriate response communicative
act ($B_{7}$), and to the best of the $Initiator$'s knowledge the interaction with $Participant$ has not yet
begun ($P_{1}$). The $Broker$ agent can send one of two messages, either a \emph{failure-proxy} or an
\emph{inform-done-proxy}, depending on whether it has succeeded to send the \emph{proxied-communicative-act}
message to $Participant$. The \emph{failure-proxy} message causes the agents to terminate the interaction with
corresponding $Participant$ agent and to transition to $I_{6}B_{8}P_{1}$ place. This place denotes a joint
interaction state in which $Initiator$ has received a \emph{failure-proxy} communicative act and terminated
($I_{6}$), $Broker$ has sent the \emph{failure-proxy} message and terminated as well ($B_{8}$) and the
interaction with the $Participant$ agent has never started ($P_{1}$). On the other hand, the
\emph{inform-done-proxy} causes the agents to transition to $I_{7}B_{9}P_{2}$ place. The $I_{7}B_{9}P_{2}$ place
represents an interaction state where $Broker$ has sent the \emph{inform-done-proxy} message ($B_{9}$),
$Initiator$ has received it ($I_{7}$), and $Participant$ has begun the interaction with the $Broker$ agent
($P_{2}$). Again, this is represented using the XOR-decision building block.

Finally, the $Broker$ agent can either send a \emph{reply-message-sub-protocol} or
 a \emph{failure-brokering} communicative act. The
\emph{failure-brokering} message causes the interacting agents to transition to $I_{8}B_{11}P_{2}$ place. This
place indicates that $Broker$ has sent a \emph{failure-brokering} message and terminated ($B_{11}$), $Initiator$
has received the message and terminated ($I_{8}$), and $Participant$ has terminated during the interaction with
the $Broker$ agent ($P_{2}$). The \emph{reply-message-sub-protocol} communicative act causes the agents to
transition to $I_{9}B_{12}P_{3}$ place. The $I_{9}B_{12}P_{3}$ place indicates that $Broker$ has sent a
\emph{reply-message-sub-protocol} message and terminated ($B_{12}$), $Initiator$ has received the message and
terminated ($I_{9}$), and $Participant$ has successfully completed the nested \emph{sub-protocol} with the
$Broker$ agent and terminated as well ($P_{3}$). Thus, the $B_{10}P_{3}$ place, denoting a successful completion
of the nested \emph{sub-protocol}, is also the corresponding transition input place.

\bibliography{AUML2CP-nets}
\bibliographystyle{theapa}

\end{document}